\documentclass{article}

\usepackage[eandd, final]{neurips_2026}
\usepackage[utf8]{inputenc} % allow utf-8 input
\usepackage[T1]{fontenc}    % use 8-bit T1 fonts
\usepackage{hyperref}       % hyperlinks
\usepackage{url}            % simple URL typesetting
\usepackage{booktabs}       % professional-quality tables
\usepackage{amsfonts}       % blackboard math symbols
\usepackage{nicefrac}       % compact symbols for 1/2, etc.
\usepackage{microtype}      % microtypography
\usepackage{xcolor}         % colors

\usepackage{graphicx}
\usepackage{multicol}
\usepackage{multirow}
\usepackage{subcaption}
\usepackage{comment}
\usepackage{placeins}
\usepackage{amsmath}

\usepackage{amssymb} % For \checkmark and \texttimes
\newcommand{\cmark}{\checkmark}
\newcommand{\xmark}{\texttimes}

\newcommand{\AB}[1]{{\color{red}AB: #1}}

\title{A$^3$Bench: Compositional Reasoning over Aggressive Interactions in Videos}

\author{%
  Austin C Baker\thanks{Equal Contribution \qquad \ensuremath{\dagger} Intern at IAI, UCF},
  Raiyaan Abdullah\footnotemark[1],
  Daniel Zeev Yoffe\textsuperscript{\ensuremath{\dagger}},
  Shruti Vyas,
  Yogesh S. Rawat \\[3pt]
  Institute of Artificial Intelligence, University of Central Florida \\[3pt]
  \texttt{au182598@ucf.edu},
  \texttt{raiyaanabdullah@gmail.com} \\
  \texttt{daniel.yoffe09@gmail.com},
  \texttt{shruti@ucf.edu},
  \texttt{yogesh@crcv.ucf.edu} \\[3pt]
  \normalfont Project page: \url{https://a3bench.github.io/a3bench/}
}

\begin{document}

\maketitle

\begin{abstract}

Recent Multimodal Large Language Models (MLLMs) have achieved strong performance on general video understanding benchmarks, yet their ability to reason about participant roles in aggressive interactions remains largely unexplored. Existing violence and crime datasets primarily evaluate whether aggression occurs, without testing whether models can identify who performed the action, who received it, and how participants are related within the scene. To address this gap, we introduce \textbf{A$^3$Bench}, a benchmark to study \textit{aggressive action and association} in videos. The benchmark evaluates whether models can jointly recognize aggressive actions, distinguish aggressors from victims and bystanders, and correctly bind actions to participants under challenging distractor settings.
We benchmark 17 multimodal models and find that they struggle substantially on fine-grained aggressive-scene understanding. Performance degrades consistently as reasoning complexity increases, with models frequently confusing aggressors with victims or bystanders, revealing a systematic failure in role-action association rather than coarse aggression recognition alone. We further observe that increasing model scale or generic reasoning models experience limited improvement. Finally, we explore a simple role-graph prompting strategy that encourages structured reasoning over participant interactions, leading to improvements in compositional reasoning. Our findings highlight that reliable aggressive-scene understanding remains a major challenge for current MLLMs and motivate future research on role-aware video reasoning beyond coarse violence detection. Code and dataset: \url{https://github.com/austincbaker/A3Bench}.
  
% To address this gap, we propose a progressive fine-tuning pipeline: supervised fine-tuning teaches MCQ format compliance, chain-of-thought distillation adds structured reasoning for compositional questions, and anchored direct preference optimization sharpens discrimination against taxonomy-hard distractors.
% Training on just 20\% of videos with group-aware splits that prevent
% parent-video leakage, our pipeline improves accuracy to \textbf{XX\%}
% (+XX points), with the largest gains on the compositional categories
% that zero-shot models struggle with most.
\end{abstract}    
\begin{figure}[t!]
  \centering
    \centering
    \includegraphics[width=\linewidth]{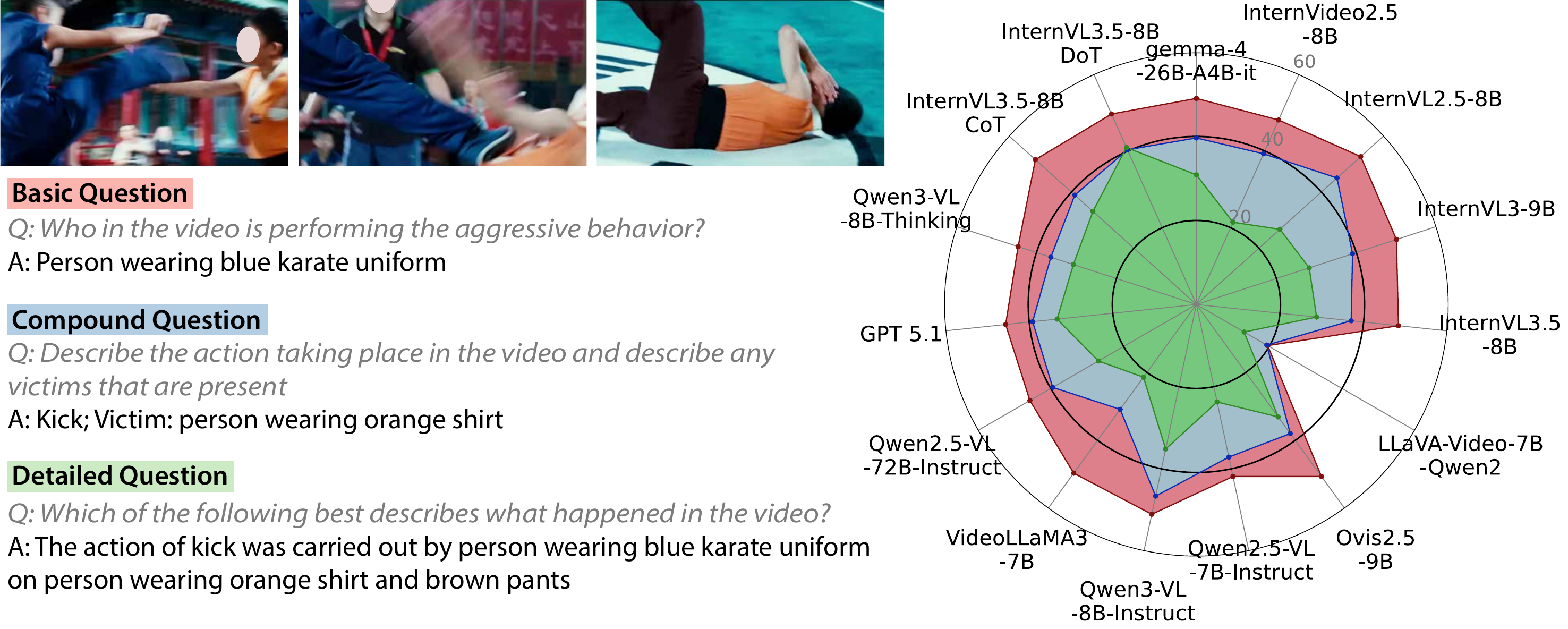}
  \caption{\textbf{A$^3$Bench evaluates fine-grained aggressive action and association.}
\textit{Left:} Example question types. The benchmark progresses from basic questions that ask about a single attribute, to detailed questions that require understanding all aspects. \textit{Right:} Performance of evaluated MLLMs across the three question levels. Models perform poorly overall scoring below 50\%, and accuracy decreases as the task becomes more difficult, with the lowest scores on detailed questions. 
% This highlights the difficulty of moving from coarse aggression recognition to reliable actor-role and action binding.
}
  \label{fig:teaser}
\end{figure}

\section{Introduction}

Recent Multimodal Large Language Models (MLLMs) have demonstrated strong performance on video understanding tasks, including action recognition \cite{kinetics, ssv2, ucf101, hmdb51, activitynet}, temporal event localization, and video question answering \cite{mvbench, tempcompass, mmmu}. Existing action datasets have been highly effective in evaluating models across a wide range of everyday activities, sports actions \cite{multisports, finegym}, human-object interactions \cite{charades, actiongenome}, and complex temporal events \cite{momentsintime, multithumos}. While these tasks are important for general video understanding, reliable real-world systems must also recognize safety-critical behaviors, including physical altercations or threatening interactions. However, understanding aggression requires more than detecting that a violent or harmful action has occurred. In practical settings such as security, public safety monitoring, school safety, sports analysis, and online video moderation, it is often not enough to recognize what is happening; it is also necessary to understand how each person in the scene is involved. Without role-aware reasoning, a system may correctly classify a scene as aggressive while still producing an incorrect or harmful interpretation.

Prior work has studied aggressive actions through crime detection \cite{ucfcrimes, crxk}, violence detection \cite{xdviolence, cabad, moviesfight, violentflows, rwf, dvd, vid}, and incident understanding datasets \cite{iard, shanghaitech}. Some provide multiple aggressive action categories, while others formulate the task as binary classification, such as normal vs. anomalous or violent vs. non-violent. These resources have advanced the study of abnormal and dangerous events, but they typically evaluate only whether aggression is present. They often do not require reasoning about the individuals involved in the scene, including who initiates the action, or who is affected by it. In particular, existing evaluations rarely test whether an aggressive action can be bound to the correct actor, whether aggressors can be distinguished from victims and bystanders, or whether multi-person interactions can be resolved. This limitation can reward models that rely on contextual shortcuts, scene priors, or coarse motion cues, rather than learning the semantic relationships between people, actions, and roles. This necessitates a more fine-grained capability: \textbf{aggressive action and association}, where the goal is not only to recognize an aggressive behavior, but also to associate it with the correct participant roles.

% To address this, we introduce A$^3$Bench, a benchmark for evaluating aggressive action and association in MLLMs. This contains 2,683 video clips and 15,004 video question-answer pairs designed to probe models beyond coarse aggression recognition. Instead of asking only whether a violent action occurs, our benchmark asks detailed questions about the type of aggressive action, and action-role relationships (aggressors, victims, bystanders). These questions require models to interpret the scene composition, distinguish the appearance between individuals, recognize and bind aggressive actions to the correct participants. By evaluating these fine-grained reasoning abilities, A$^3$Bench provides a focused testbed for assessing whether current MLLMs can reliably understand aggressive scenes at the level required for real-world deployment. 

To address this, we introduce A$^3$Bench, a benchmark for evaluating aggressive action and association (A$^3$) in MLLMs. A$^3$Bench contains 2,683 video clips and 15,004 question-answer pairs that move beyond coarse aggression recognition to evaluate role-aware reasoning in aggressive scenes. Rather than asking only whether aggression occurs, the benchmark probes models on aggressive actions and participant-role associations, including aggressors, victims, and bystanders. These questions require models to distinguish individuals, interpret interactions, and correctly bind actions to participants. By focusing on fine-grained role-action reasoning, A$^3$Bench provides a challenging testbed for assessing whether current MLLMs can reliably understand aggressive scenes.

% Our experiments show that recent MLLMs struggle with this evaluation with models achieving performance below 50\%. Our analysis shows that model performance drops as questions become more compositional, decreasing from questions that query a single attribute to questions that require full scene reasoning (Figure~\ref{fig:teaser}, right). Error analysis further reveals that models often identify visually salient participants without recognizing the correct aggressive action, and tend to over-assign the aggressor role to victims or bystanders. These failures suggest that current MLLMs do not reliably follow the interaction structure needed for aggressive-scene understanding. To explore whether explicit structure can help, we introduce a role-graph prompting strategy, which decomposes the scene into structured role-action reasoning. This encourages models to first determine who is physically involved, then distinguish who initiates vs. receives the action, and only then infer the aggressive action.

Our experiments show that recent MLLMs struggle on this benchmark, with performance remaining below 50\%. Accuracy consistently drops as questions require more compositional reasoning, from identifying a single attribute to jointly reasoning about actions and participant roles (Figure~\ref{fig:teaser}, right). Error analysis further reveals a strong bias toward visually salient individuals, with models frequently misclassifying victims or bystanders as aggressors. These findings suggest that current MLLMs fail to reliably capture the interaction structure underlying aggressive scenes. Finally, we explore a simple role-graph prompting strategy that encourages structured reasoning over participant interactions, leading to improved performance on compositional reasoning tasks.

\iffalse
Overall, our contributions are as follows:
\begin{itemize}
    
    \item We introduce A$^3$Bench, a benchmark of 2,683 video clips designed to evaluate fine-grained aggressive behavior understanding in MLLMs.
    
    \item We construct diverse question types of 15,004 video question-answer pairs for deep understanding by probing to role identification along with aggressive action understanding.
    
    \item We benchmark recent MLLMs revealing limitations in their ability to properly understand aggressive scenes.

    \item We propose a role-graph prompting strategy that explicitly decomposes aggressive-scene understanding into structured role-action reasoning to guide MLLMs.
\end{itemize}
\fi

We make the following contributions: 1) We introduce A$^3$Bench, a benchmark for evaluating aggressive action and association in videos designed to assess fine-grained role-aware reasoning beyond coarse violence recognition, 2) We conduct a comprehensive evaluation of recent multimodal large language models in a zero-shot setting, showing that current systems struggle to reliably associate aggressive actions with the correct participants, with performance degrading significantly as reasoning complexity increases, and 3) Through detailed analysis, we identify systematic failures in role-action association, and show that a simple role-graph prompting strategy can improve structured reasoning in aggressive-scene understanding. 

\begin{comment}
\AB{Potential insights?:
(a) Aggressor bias: models systematically over-assign the aggressor role -- 37\% of true victims and 31\% of bystanders are mislabeled as aggressors, meaning a deployed system would routinely blame victims.
(b) Scale does not help: Qwen2.5-VL-72B (40.53\%) scores below several 8B models like Qwen3-VL-8B (46.96\%) and InternVL2.5-8B (45.77\%). A 10x parameter increase does not buy compositional aggression understanding.
(c) People without action: models identify the correct participant 30\% of the time while getting the action wrong, but the reverse (correct action, wrong person) is only 18\%. They find salient people but cannot determine what the aggression actually is.
(d) Reasoning is a dead end: thinking modes and CoT prompts average only 42.83\% overall, failing to outperform the best general-purpose model. Step-by-step reasoning does not fix the grounding deficit.
Recommendation: (a) + (b) together. aggressor bias is the most novel finding with real-world safety implications, and the scale result preempts the obvious ``just use a bigger model'' counterargument.}
\end{comment}
\section{Related Work}
\noindent\textbf{Action recognition:}
The advancement of action recognition can be noted across video \cite{VideoMAE, ActionDetailLAQ}, skeleton \cite{FreqSemSkeleton, CrossModalSkeleton, skeletoncontext}, and multimodal modalities \cite{vificlip, actionclip, xclip, froster, t2l, AudioVisualLearning}. Certain methods, such as temporal localization \cite{TriDet, Stepwise, PseudoFormer} allow more precise event boundary detection. Fine-grained recognition has advanced through unified keypoint-based frameworks \cite{Keypoint} and atomic action decomposition \cite{Atomic}. Action detection methods further localize and identify what each person is doing in space and time, but they typically focus on individual actions rather than the relationships between interacting participants. These methods still treat actions as a classification problem over participant-agnostic categories, without modeling the relationships between those involved.

\noindent\textbf{Crime and anomaly detection:}
Anomaly detection and behavior analysis in security videos have been widely studied for identifying abnormal, unsafe, or criminal activities~\cite{OSAR,surveillancevqa}. In parallel, dedicated violence and aggression datasets have advanced the evaluation of models across diverse harmful scenarios, including fight detection, crowd violence, real-world security violence, intruder activity, child aggression, and crime/anomaly recognition~\cite{moviesfight,dvd,rlvs,violentflows,rwf,vid,iard,xdviolence,cabad,crxk,ucfcrimes}. Beyond visual understanding, text-based aggression detection has also examined aggressive behavior in language, such as harmful or hostile social-media comments~\cite{AggressiveComments}, showing that aggression analysis spans multiple modalities. Most existing video benchmarks focus on violence, aggression, or anomaly detection with coarse event-level labels. In contrast, A$^3$Bench evaluates fine-grained aggressive action association by requiring models to identify the actor, target, and role of each visible participant, exposing failures that prior benchmarks do not measure.

\noindent\textbf{Compositional video reasoning:}
Existing benchmarks have explored compositional video understanding through temporal and causal reasoning~\cite{cleverer}, spatio-temporal question answering~\cite{agqa}, and situated reasoning over human interactions~\cite{star}. More recent work evaluates fine-grained associations between agents, actions, and events using video-language entailment and challenging negative captions~\cite{velociti}. While these assess general compositional reasoning, A$^3$Bench focuses on aggressive interactions, requiring models to distinguish aggressors, victims, and bystanders and correctly associate actions with participant roles and interaction direction.

\noindent\textbf{Multimodal Large Language Models (MLLMs):}
Recent multimodal models have significantly expanded the scope of visual understanding by linking visual evidence with language-based semantic reasoning~\cite{clip}. Building on large-scale cross-modal pretraining, such models have enabled zero-shot and open-vocabulary recognition of human actions and activities~\cite{VLMeffective,actionclip,xclip,vificlip,froster,t2l,internvid}. Recent video-centric MLLMs~\cite{qwen3vl,internvl3,ovis2_5,gemma4}, further extend these capabilities to temporal and multi-frame reasoning over complex events~\cite{VideoGPT,SeViLA,LVR}. These advances motivate the use of MLLMs for aggressive action and association, where models must not only identify the observed action but also reason about the relationships between people and their roles.

\section{Aggressive Action and Association Benchmark (A\texorpdfstring{$^3$}{³}Bench)}
A$^3$Bench is designed to evaluate fine-grained aggressive action and association in real-world videos. The benchmark construction begins with video curation and structured annotation of aggressive interactions, including actions, participants, roles, and scene context. These annotations are then converted into three levels of diagnostic VQA questions, ranging from basic attribute recognition to compound association and detailed aggressor-action-victim reasoning. The resulting benchmark enables controlled evaluation of MLLMs under a unified multiple-choice protocol.

\begin{table}[t!]
    \small
    \centering
    \caption{\textbf{Comparison of A$^3$Bench with prior video benchmarks spanning aggression recognition and compositional reasoning.}
    Desc. denotes natural-language person descriptions; Aggression Roles indicates explicit aggressor, victim, and bystander annotations; VQA/Caption Pairs reports question-answer or video-caption pairs; Relational Structure denotes the relational or role annotation format; and Task Format specifies the evaluation protocol.}
    \label{tab:benchmark_comparison}
    \resizebox{\textwidth}{!}{%
    \begin{tabular}{lrrccclll}
        \toprule
        \textbf{Benchmark} & \textbf{Actions/} & \textbf{\# Videos} & \textbf{Desc.} & \textbf{Aggression} & \textbf{Aggression} & \textbf{VQA/Caption} & \textbf{Relational} & \textbf{Task} \\
        \textbf{} & \textbf{Verbs} & \textbf{} & \textbf{} & \textbf{} & \textbf{Roles} & \textbf{Pairs} & \textbf{Structure} & \textbf{Format} \\
        \midrule
        DVD \cite{dvd}                  & 2   & 500    & \xmark & \cmark & \xmark & N/A  & N/A                          & Classification \\
        Movies Fight \cite{moviesfight} & 2   & 200    & \xmark & \cmark & \xmark & N/A  & N/A                          & Classification \\
        RLVS \cite{rlvs}                & 2   & 2,000  & \xmark & \cmark & \xmark & N/A  & N/A                          & Classification \\
        RWF-2000 \cite{rwf}             & 2   & 2,000  & \xmark & \cmark & \xmark & N/A  & N/A                          & Classification \\
        VID \cite{vid}                  & 2   & 3,020  & \xmark & \cmark & \xmark & N/A  & N/A                          & Classification \\
        Violent Flows \cite{violentflows}       & 2   & 246    & \xmark & \cmark & \xmark & N/A  & N/A                          & Classification \\
        \midrule
        CABAD \cite{cabad}              & 6   & 900    & \xmark & \cmark & \xmark & N/A  & N/A                          & Classification \\
        CRxK \cite{crxk}                & 13  & 3,484  & \xmark & \cmark & \xmark & N/A  & N/A                          & Classification \\
        IARD \cite{iard}                & 4   & 802    & \xmark & \cmark & \xmark & N/A  & N/A                          & Classification \\
        UCF-Crime \cite{ucfcrimes}      & 14  & 1,900  & \xmark & \cmark & \xmark & N/A  & N/A                          & Classification \\
        XD-Violence \cite{xdviolence}   & 7   & 4,754  & \xmark & \cmark & \xmark & N/A  & N/A                          & Classification \\
        \midrule
        CLEVRER \cite{cleverer}         & --  & 20,000 & \xmark & \xmark & \xmark & 300K & Programmatic                 & Open QA \\
        AGQA \cite{agqa}                & 157 & 9,600  & \xmark & \xmark & \xmark & 192M & Scene Graph                  & Open QA / MCQ \\
        STAR \cite{star}                & 111 & 9,000  & \xmark & \xmark & \xmark & 60K  & Situation Graph              & MCQ \\
        VELOCITI \cite{velociti}        & --  & 864    & \cmark & \xmark & \xmark & 3K   & Semantic Role Label          & Binary Entailment \\
        \midrule
        \textbf{A$^3$Bench (Ours)}      & \textbf{18} & \textbf{2,683} & \cmark & \cmark & \cmark & \textbf{15K} & \textbf{Structured} & \textbf{MCQ} \\
             & \textbf{} & \textbf{} &  &  &  &  & \textbf{Participant Roles} & \\
        \bottomrule
    \end{tabular}%
    }
\end{table}

\subsection{Construction}
\textbf{Video curation and annotation:}
We curate a benchmark of 2,683 video clips collected from diverse sources, including publicly available social-media videos, UCF-Crime~\cite{ucfcrimes}, and RWF-2000~\cite{rwf}. Source videos are trimmed into shorter clips centered on individual interactions, resulting in an average clip duration of 5.75 seconds (standard deviation of 1.69 seconds), with most clips approximately 4 to 8 seconds long. The videos span a wide range of recording conditions, camera viewpoints, resolutions, scene layouts, and levels of visual clutter, reflecting the heterogeneity of real-world scenarios. The final corpus contains 2,234 clips with aggressive behavior and 449 non-aggressive clips.

\begin{figure}[t]
  \centering
  \includegraphics[width=\linewidth]{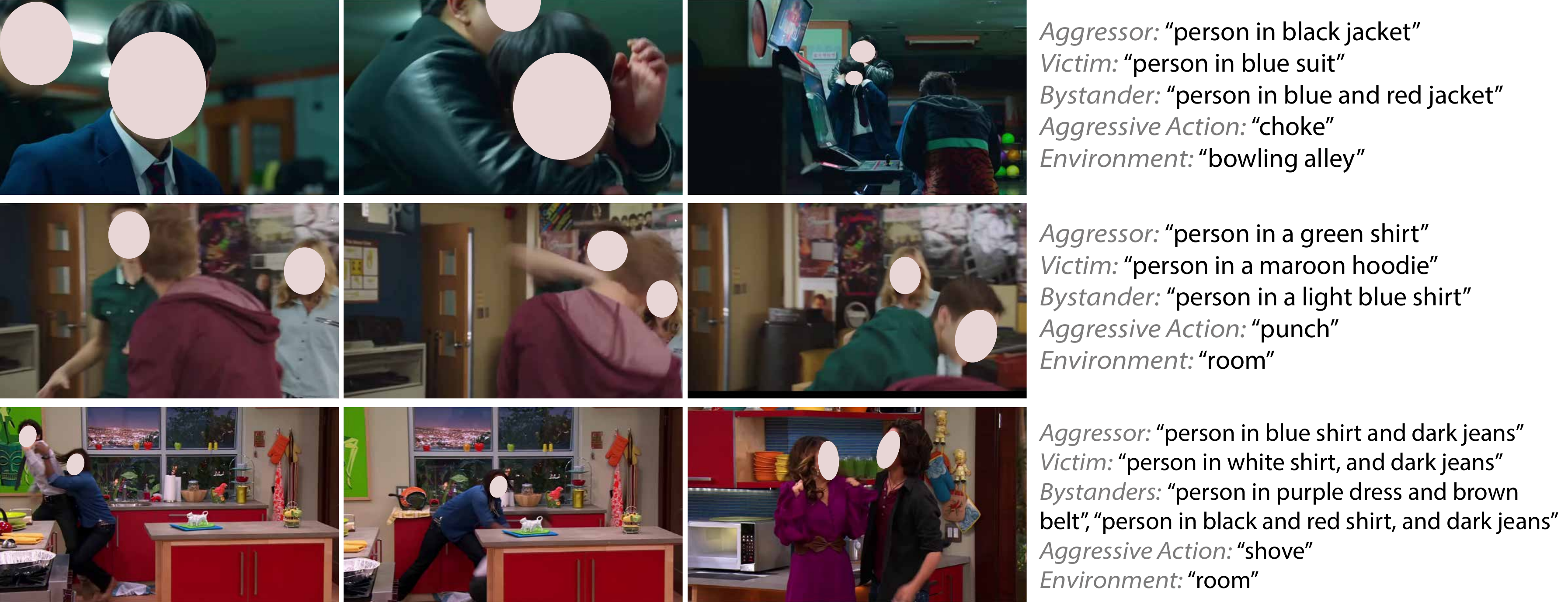}
  \caption{\textbf{Example videos from A$^3$Bench.}
Each video is annotated with structured role and scene information, including the aggressor, victim, bystanders, aggressive action, and environment. Person descriptions are based primarily on clothing and coarse visual attributes rather than identity, enabling fine-grained actor-role association while avoiding personally identifying information.
}
  \label{fig:dataset_video_examples}
\end{figure}

A group of annotators labeled each video with structured scene information. The annotations include: (1) the aggressive action being performed, selected from 17 aggressive action categories and a control ``none'' category for non-aggressive clips; a short appearance description of the (2) aggressor; (3) victim; (4) bystanders and (5) the environment or setting. As seen in Figure~\ref{fig:dataset_video_examples}, person descriptions are primarily based on clothing and coarse visual appearance, such as ``person in a dark jacket,'' while avoiding names, faces, demographic attributes, or other identifying information. This design reduces the risk of relying on sensitive attributes and focuses the benchmark on visual role and action understanding. If a certain role does not exist in the video clip, or the appearance is too ambiguous, that is left out in the annotations. 

\begin{figure}[t!]
  \centering
    \centering
    \includegraphics[width=\linewidth]{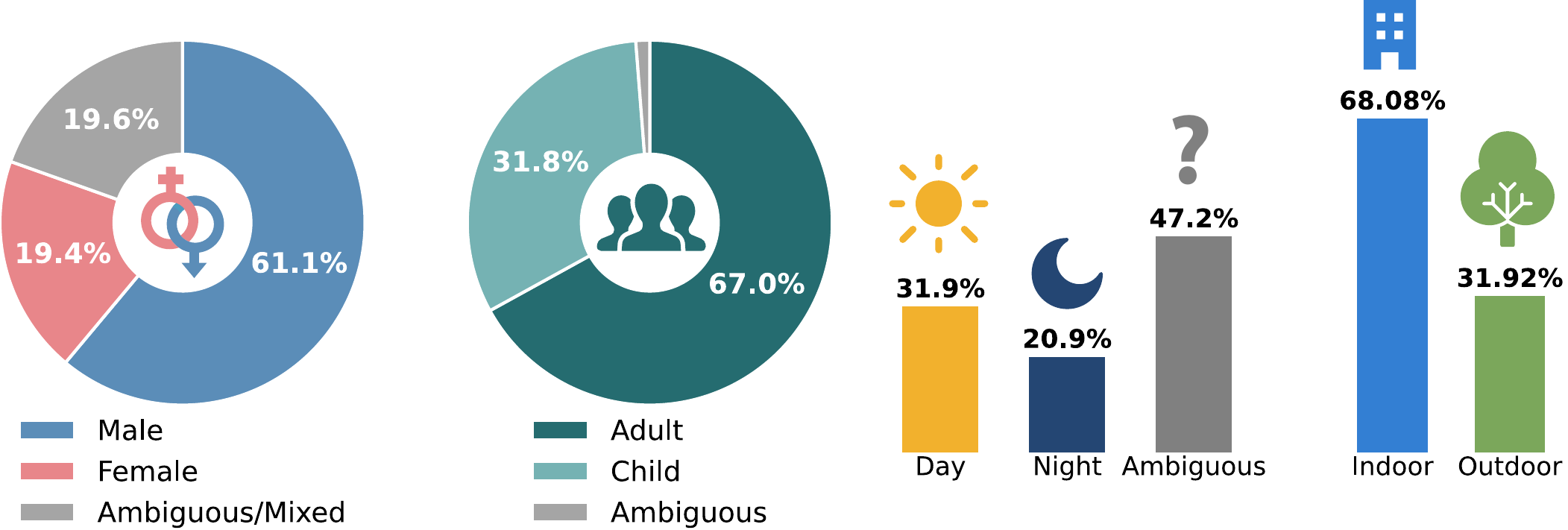}
  \caption{\textbf{Demography (aggressors/victims) and environment settings.} A$^3$Bench contains a larger proportion of male and adult participants (aggressors or victims) among videos where demographic attributes are identifiable. For scene conditions, many clips are ambiguous in time of day, but among identifiable cases, slightly more aggressive clips occur during daytime than night, and most are captured in indoor environments.
}
  \label{fig:dataset_demograph}
\end{figure}

\noindent\textbf{Dataset distribution.}
A$^3$Bench intentionally contains more aggressive than non-aggressive clips, as its primary objective is to evaluate fine-grained action and participant-role reasoning within aggressive interactions. Non-aggressive clips are included to evaluate whether models can recognize the absence of aggression. As shown in Figure~\ref{fig:dataset_demograph}, A$^3$Bench also exhibits a male-skewed participant distribution, reflecting the composition of the collected real-world source videos. This observation is consistent with broader crime statistics reporting higher male representation in several offense categories. According to the FBI's 2019 arrest statistics, males accounted for 72.5\% of all arrests, including 84.2\% of robbery arrests, and 76.5\% of aggravated-assault arrests~\cite{fbi2019}. UNODC Global Study on Homicide 2023 also reports that approximately 91\% of suspects brought to police for intentional homicide (across 50 countries) between 2018 and 2021 were male~\cite{unodc2023}.

\noindent\textbf{Annotation quality assurance:}
We employ a structured, consensus-driven annotation protocol involving four annotators to ensure visually grounded and reliable annotations of aggressive interactions. Each annotator independently reviews assigned videos and labels the aggressive-action category, appearance descriptions of aggressors, victims, and bystanders, and scene environment. Participants are identified using clothing and other coarse, non-identifying visual cues (e.g., ``person in a red shirt''), while avoiding race, gender, age, and other sensitive attributes. Demographic statistics in Figure~\ref{fig:dataset_demograph} are reported only at the aggregate corpus level and are not used in questions, answer options, or role labels.

Annotators flag videos with ambiguous actions, interaction direction, participant visibility, or appearance descriptions, particularly in crowded, occluded, blurry, black-and-white, or low-resolution scenes. Flagged annotations are withheld from finalization pending group review. After each annotation batch, another team member independently re-reviews a randomly selected subset. Aggressor-victim direction conflicts occur in fewer than 5\% of cross-checked cases, while appearance-description disagreements occur in approximately 20\%, primarily because annotators select different clothing or coarse visual cues to describe the same participant rather than identifying different participants. All flagged cases and disagreements identified during cross-checking are collectively reviewed. Role-direction disagreements are resolved by determining who initiated and who received the aggressive action, while appearance-description disagreements are resolved by selecting the clearest non-identifying description. When reliable consensus cannot be reached, the corresponding attribute is omitted rather than guessed. Periodic spot checks and targeted re-review of difficult cases continue throughout benchmark development, refining participant descriptions and removing ambiguous attributes while retaining only visually supported annotations.

\begin{figure}[h!]
  \centering
  \begin{minipage}{0.64\linewidth}
    \centering
    \includegraphics[width=\linewidth]{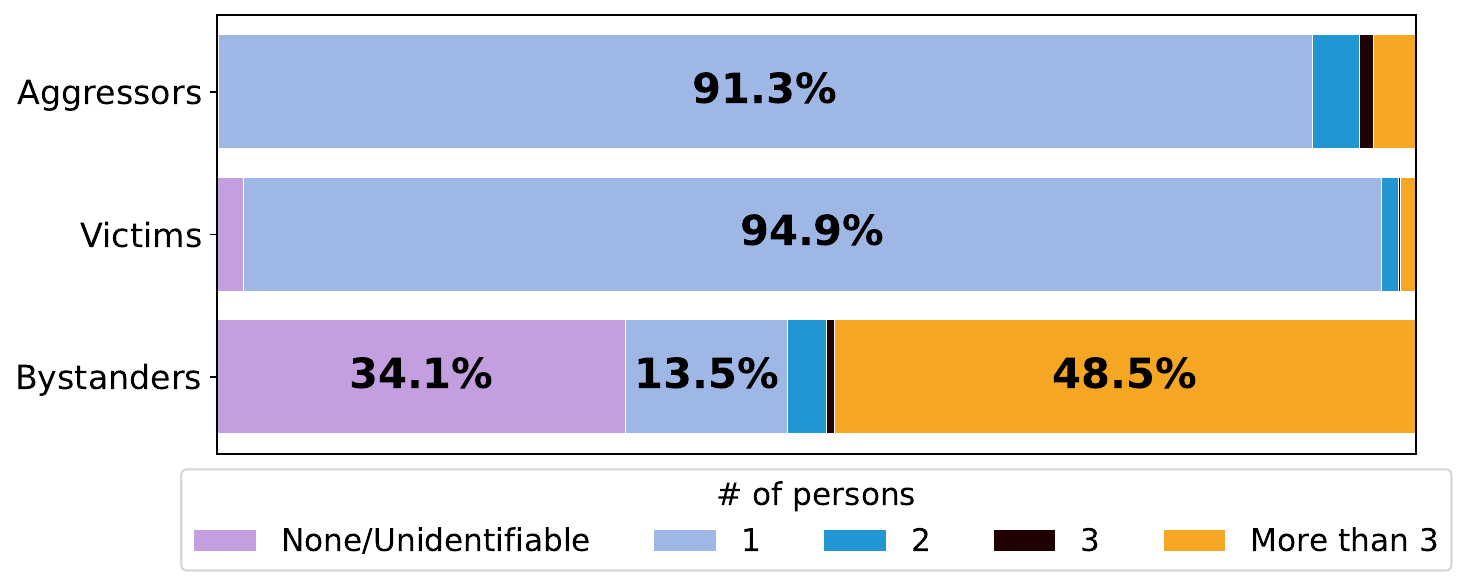}
  \end{minipage}%
  \hfill
  \begin{minipage}{0.34\linewidth}
    \captionof{figure}{\textbf{Distribution of annotated roles per video.}
The role-count distribution is also skewed towards most clips containing a single identifiable aggressor and victim, while bystanders are more variable and often appear in larger groups.
}
    \label{fig:video_role_count}
  \end{minipage}
\end{figure}

\begin{figure}[h!]
  \centering
  \begin{minipage}{0.64\linewidth}
    \centering
    \includegraphics[width=\linewidth]{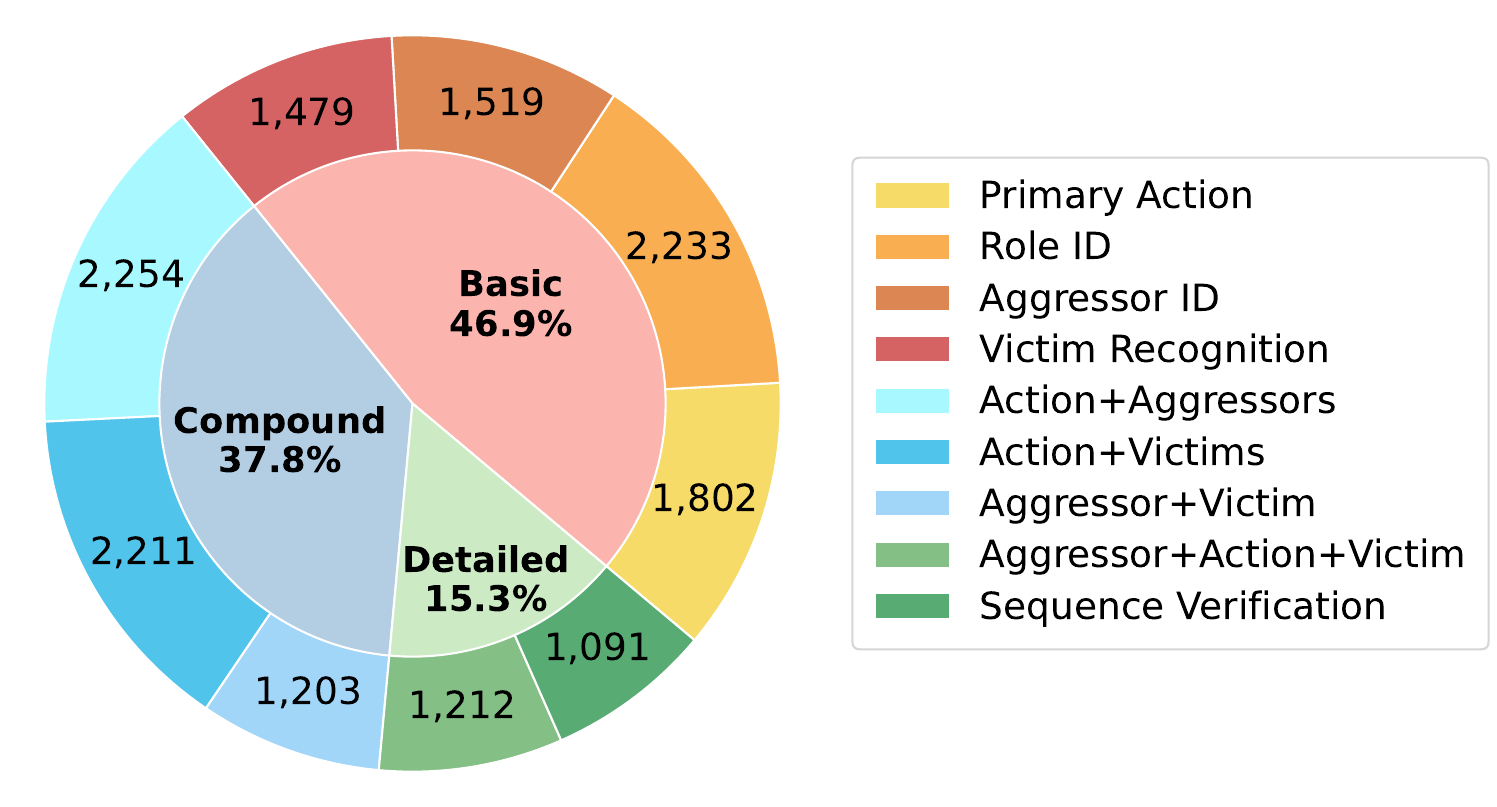}
  \end{minipage}%
  \hfill
  \begin{minipage}{0.34\linewidth}
    \captionof{figure}{\textbf{Distribution of question types in A$^3$Bench.}
The benchmark contains three levels of questions with increasing association complexity: \textit{Basic} questions focus on single attributes; \textit{Compound} questions require associating two scene components; and \textit{Detailed} questions require full interaction understanding.
}
    \label{fig:question_type_category}
  \end{minipage}
\end{figure}

\textbf{Question generation and distractor design:}
From the annotation corpus, we generate multiple-choice video question-answer pairs that function as controlled diagnostic probes for aggressive interaction understanding. Rather than only evaluating whether a model can recognize an action label, A$^3$Bench is designed to expose distinct failure modes in video-language models, including weak role grounding, poor action-participant binding, confusion between visible and relevant people, failure to model interaction directionality, and reliance on answer-choice priors.  The questions are organized into three tiers of increasing complexity: \textit{basic questions}, \textit{compound questions}, and \textit{detailed questions}. Figure~\ref{fig:teaser} illustrates representative examples from these three categories, and Figure~\ref{fig:question_type_category} reports their subcategories and overall distribution across the benchmark.

\textit{Basic questions} test whether a model can recover a single atomic attribute of the scene. These include primary action identification, where the model must recognize the aggressive action taking place; aggressor or victim identification, where the model must identify who performs the aggression or who is targeted; and role identification, where the model is given a person description and must assign the correct role, such as aggressor, victim, or bystander. These questions diagnose whether the model can localize individual actions and participant roles before requiring relational reasoning.

\textit{Compound questions} test whether a model can bind two scene attributes correctly. These include aggressor-victim, action-aggressor, and action-victim associations. Since each answer choice is only correct when all components are jointly correct, these questions diagnose whether models merely recognize visible actions or participants independently, or whether they can associate the correct action with the correct person and role.

\textit{Detailed questions} test whether a model can recover the full event structure of an aggressive interaction. These include aggressor-action-victim questions, which ask who did what to whom, and sequence verification questions, which require selecting the correct structured description of the interaction. These questions specifically diagnose failures in directionality, role distinction, and holistic event understanding, since a model must distinguish the aggressor from the victim and construct a coherent representation of the interaction rather than predict a participant-agnostic action category.

Importantly, Basic, Compound, and Detailed questions are generated from the same underlying video pool, with each video contributing approximately seven questions across multiple complexity levels. 

We design distractor choices as targeted diagnostic stress tests: they are visually plausible, semantically close to the correct answer, and tied to specific failure modes. \textit{Cross-video distractors} are sampled from real annotations of other clips, so incorrect options still contain natural actions, roles, and participant descriptions rather than synthetic or obviously invalid answers. \textit{Same-video distractors} use visible non-target participants from the queried clip, such as the victim or a bystander when asking for the aggressor, preventing models from relying only on visibility, saliency, or person presence. \textit{Role reversals and bystander substitutions} preserve much of the event content while swapping aggressor and victim roles or assigning bystanders to active roles, directly testing interaction directionality and the distinction between active participation and mere presence. \textit{Negation questions} include answers such as ``no aggressive action is taking place'' to test whether models verify visual evidence rather than assume aggression is present. \textit{Frequency-inverted distractors} are constructed so that each key distractor property, such as an action, role, or participant description, appears across multiple answer options, making marginal answer frequency uninformative. This forces models to reason over the complete aggressor-action-victim structure rather than selecting the option with the most common or familiar textual components.

\subsection{Benchmarked Models}

We evaluate \textbf{17} representative model configurations on \textbf{A$^3$Bench} to assess their ability to perform aggressive action and association. All of our chosen models are trained on very large datasets with high-level video understanding. 

\noindent \textbf{General-Purpose Models.}
The evaluated models include two CLIP-based video-language models, ViCLIP~\cite{internvid} and X-CLIP~\cite{xclip}, alongside several recent open-weight MLLM families, including Gemma \cite{gemma4}, InternVideo \cite{internvideo2_5}, InternVL \cite{internvl2_5, internvl3, internvl3_5}, LLaVA-Video \cite{llavavideo}, Ovis \cite{ovis2_5}, Qwen-VL \cite{qwen-vl2_5, qwen3vl}, and VideoLLaMA \cite{videollama3}. Among closed-source, we include GPT 5.1 \cite{gpt_51}. These families differ in visual encoder design, video modeling strategy, instruction tuning, model scale, and reasoning behavior. 

\noindent \textbf{Reasoning and Prompted Variants.}
To investigate whether explicit reasoning improves aggressive action and association, we additionally evaluate reasoning-oriented variant of Qwen3-VL-8B \cite{qwen3vl}. We also evaluate InternVL3.5-8B \cite{internvl3_5} under two reasoning-style prompting settings: a detailed chain-of-thought (CoT) prompt and a Dream-of-Thoughts (inspired from \cite{navigatinghallucinations}) prompt. These variants encourage intermediate reasoning before producing a final answer, allowing us to test whether step-by-step inference helps models distinguish aggressors, victims, bystanders, actions, and scene context.

\subsection{Implementation Details, Evaluation Protocol, and Metrics}

All experiments are zero-shot. Open-source models are evaluated using their official implementations, hyperparameters, and released weights on a single NVIDIA RTX A6000 (48 GB) or A100 (80 GB) GPU. Qwen2.5-VL-72B-Instruct \cite{qwen-vl2_5} was run on the quantized version to fit in memory. We use a multiple-choice video question answering (VQA) protocol across all models: for each video-question pair, the model is prompted to select the most appropriate answer from the provided options. Depending on the question type and the size of the valid answer space, each question contains four to eight answer choices, including one correct answer and a set of diagnostic distractors. The position of the correct answer is uniformly shuffled across all questions to mitigate positional bias. 
% This controlled evaluation setting enables fair comparison across models while directly measuring each model's ability to associate aggressive actions with the correct participants and roles. 
We report accuracy as the primary evaluation metric.

\section{Experimental Results and Analysis}
Table~\ref{tab:a3bench_results} summarizes model performance on A$^3$Bench. We report both overall accuracy, computed as the question-count-weighted average, and category average, computed as the unweighted mean across Basic, Compound, and Detailed questions. Overall, current MLLMs perform poorly on the benchmark: most models achieve below 50\% accuracy, and even the best-performing model reaches only 46.96\% overall accuracy and 44.30\% category average. These results indicate that aggressive action and association remains challenging for state-of-the-art systems. In particular, current MLLMs struggle when evaluation moves beyond coarse aggression recognition and requires identifying who is involved, what action occurs, and how participants are related.

\begin{table}[t]
\small
\centering
\caption{\textbf{Benchmarking models on A$^3$Bench.} Reported metric is accuracy (\%) for Basic Questions, Compound Questions, Detailed Questions, Overall, and Average. \textbf{Best} and \underline{second-best} performance are highlighted.
}
\label{tab:a3bench_results}
\resizebox{\textwidth}{!}{
\begin{tabular}{l | c c c c c}
\toprule
\textbf{Model}
& \textbf{Basic} \textuparrow
& \textbf{Compound} \textuparrow
& \textbf{Detailed} \textuparrow
& \textbf{Overall} \textuparrow
& \textbf{Average} \textuparrow \\
\midrule
ViCLIP \cite{internvid} & 28.11 & 21.61 & 16.16 & 23.82 & 21.96\\
X-CLIP \cite{xclip} & 23.67 & 15.75 & 20.07 & 20.12 & 19.83\\
gemma-4-26B-A4B-it \cite{gemma4} & 49.07 & 39.65 & 30.86 & 42.72 & 39.86 \\
InternVideo2.5-8B \cite{internvideo2_5} & 48.07 & 39.34 & 21.51 & 40.70 & 36.31 \\
InternVL2.5-8B \cite{internvl2_5} & \textbf{52.61} & \underline{45.01} & 26.75 & \underline{45.77} & 41.46 \\
InternVL3-9B \cite{internvl3} & 50.05 & 39.06 & 28.22 & 42.55 & 39.11 \\
InternVL3.5-8B \cite{internvl3_5} & 48.34 & 37.03 & 28.75 & 41.06 & 38.04 \\
LLaVA-Video-7B-Qwen2 \cite{llavavideo} & 19.58 & 19.34 & 13.11 & 18.50 & 17.34 \\
Ovis2.5-9B \cite{ovis2_5} & 50.62 & 37.99 & 33.09 & 43.16 & 40.57 \\
Qwen2.5-VL-7B-Instruct \cite{qwen-vl2_5} & 41.87 & 37.15 & 23.73 & 37.30 & 34.25 \\
Qwen3-VL-8B-Instruct \cite{qwen3vl} & 51.07 & \textbf{46.65} & \underline{35.19} & \textbf{46.96} & \textbf{44.30} \\
VideoLLaMA3-7B \cite{videollama3} & 49.65 & 30.86 & 21.41 & 38.22 & 33.97 \\
Qwen2.5-VL-72B-Instruct \cite{qwen-vl2_5} & 45.73 & 39.47 & 26.91 & 40.53 & 37.37 \\
GPT 5.1 \cite{gpt_51} & 45.63 & 39.17 & 33.32 & 41.30 & 39.37 \\
\midrule
Qwen3-VL-8B-Thinking \cite{qwen3vl} & 44.59 & 36.41 & 30.77 & 39.37 & 37.26 \\
InternVL3.5-8B CoT prompt & \underline{51.53} & 38.92 & 33.13 & 43.94 & 41.19 \\
InternVL3.5-8B DoT \cite{navigatinghallucinations} & 49.64 & 40.31 & \textbf{40.90} & 44.77 & \underline{43.62} \\
\bottomrule
\end{tabular}
}
\end{table}

\noindent\textbf{Human evaluation.}
We additionally conduct a human evaluation involving 34 participants using the same video-question pairs and answer choices presented to the models. Each participant answers five questions from five distinct action classes, with one video sampled per class. In total, we collect 170 responses, comprising 85 Basic, 49 Compound, and 36 Detailed questions. Human participants achieve an overall accuracy of 83.5\%, highlighting substantial room for improvement of model performance in compositional understanding of aggressive interactions.

\subsection{Analysis}
\textbf{\textit{Performance degrades as aggression understanding becomes more compositional.}}
Models achieve an average accuracy of 44.11\% on \textit{Basic} questions, but performance drops to 35.51\% on \textit{Compound} questions and further to 27.29\% on \textit{Detailed} questions. This monotonic decline shows that current MLLMs do not simply fail at rare aggression labels; they fail progressively as the task requires role-action binding. Models can sometimes identify isolated visual cues, such as an action category, a salient participant, or a single role, but struggle once these cues must be composed into a structured interaction. \textit{Compound} questions expose failures in binding two attributes, such as linking the correct action to the correct aggressor or victim. \textit{Detailed} questions further require the full aggressor-action-victim structure, or who did what to whom. The sharp drop on these questions indicates failures in directionality, role assignment, and person-action binding.

Model-level trends show that the compositional drop is not uniform across systems. VideoLLaMA3-7B exhibits the steepest collapse, suggesting strong reliance on single-attribute cues. Qwen2.5-VL-7B-Instruct shows a small Basic-to-Compound drop but a much larger Compound-to-Detailed drop, indicating that pairwise association does not transfer to full aggressor-action-victim reasoning. Ovis2.5-9B shows the reverse pattern, with larger early drops and smaller later drops, suggesting that role-action binding is their main bottleneck. Qwen3-VL-8B-Instruct achieves the strongest Compound performance but still degrades on Detailed questions, while ViCLIP and LLaVA-Video-7B-Qwen2 remain weak across all tiers. The InternVL3.5-8B DoT \cite{internvl3_5, navigatinghallucinations} variant and X-CLIP are notable exceptions, with Detailed accuracy slightly exceeding Compound accuracy. 

%\AB{Model-wise, the drop-off rate varies dramatically. VideoLLaMA3-7B has the steepest collapse, suggesting it relies almost entirely on single-frame cues. InternVL2.5-78B degrades most gracefully in absolute terms (60.8\% to 45.8\%, -15pp) but within the InternVL2.5 family, the scale gap widens with complexity: +8.1pp at Basic but +19.0pp at Detailed, showing scale disproportionately helps compositional reasoning. In contrast, Qwen2.5-VL shows only a uniform +3pp gain from 7B to 72B across all tiers. Ovis2.5-9B is notable for holding up relatively well on Detailed (33.1\%) despite middling Basic scores (50.6\%), suggesting a different internal reasoning strategy.}

\textbf{\textit{Reasoning helps, but not uniformly.}}
Qwen3-VL-8B-Thinking underperforms its instruction-tuned counterpart across all categories, dropping from 46.96\% to 39.37\% overall, suggesting that generic thinking mode does not necessarily improve role-action association. In contrast, reasoning prompts improve InternVL3.5-8B: CoT and DoT increases overall accuracy. The gains are most pronounced on \textit{Detailed} questions, where InternVL3.5-8B rises from 28.75\% to 33.13\% with CoT and 40.90\% with DoT. This suggests that structured reasoning is more useful when the task requires full aggressor-action-victim binding.

%\AB{The one exception is InternVL3.5-8B-DoT (Dream of Thoughts), which achieves the flattest compositional curve of any model (49.6\% Basic to 40.9\% Detailed, only -8.7pp drop) and the second-best Detailed score (40.9\%) despite a mediocre Basic score. This is the only reasoning variant that clearly helps on the hardest questions. Qwen3-VL-8B-Thinking actually hurts: it scores 44.6\% Basic vs 51.1\% for non-thinking Qwen3, a 6.5pp regression. Ovis-Thinking is statistically identical to non-thinking Ovis (+0.08pp overall). The pattern suggests that generic ``think step-by-step'' prompting is insufficient -- the reasoning must be structured around the specific decomposition the task requires (as DoT does with its scene graph approach).}

%% 

\begin{figure}[t]
  \centering

  \begin{minipage}[t]{0.27\linewidth}
    \vspace{0pt}
    \centering
    \includegraphics[width=\linewidth]{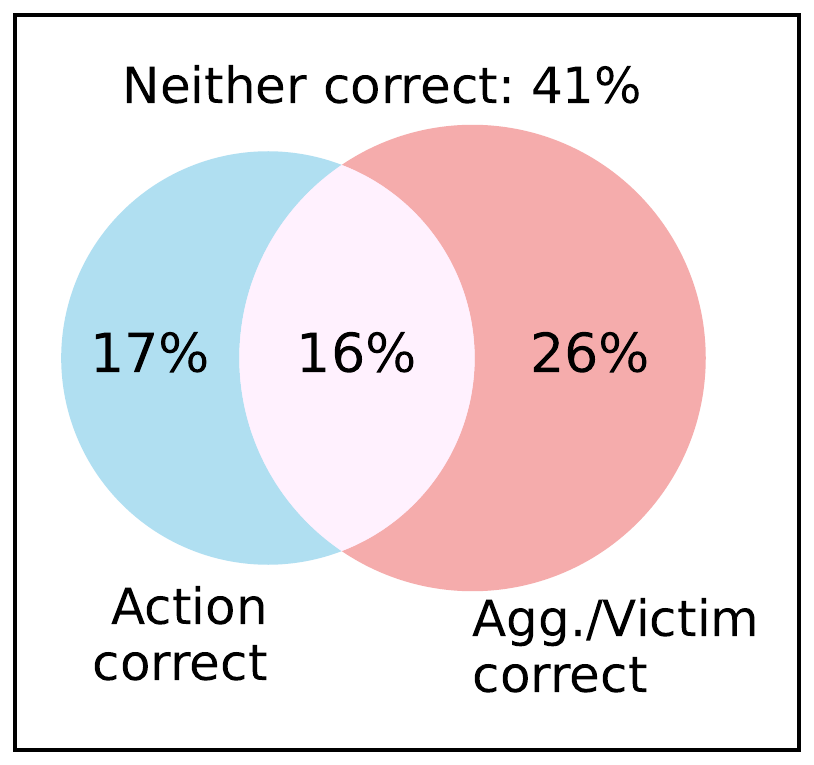}
  \end{minipage}
  \hfill
  \begin{minipage}[t]{0.27\linewidth}
    \vspace{0pt}
    \centering
    \includegraphics[width=\linewidth]{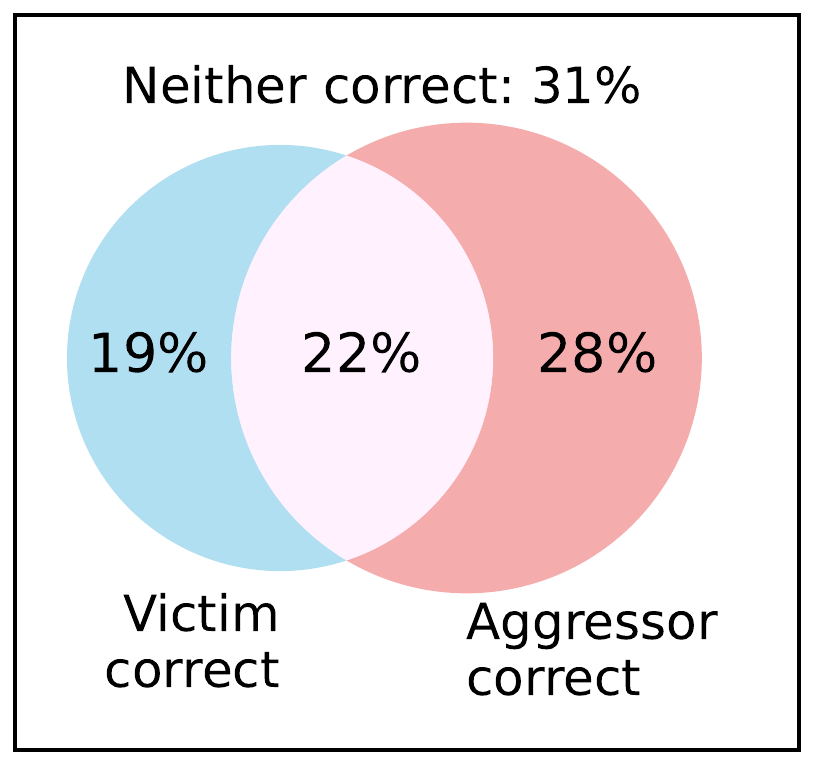}
  \end{minipage}
  \hfill
  \begin{minipage}[t]{0.43\linewidth}
    \vspace{0pt}
    \centering
    \includegraphics[width=\linewidth]{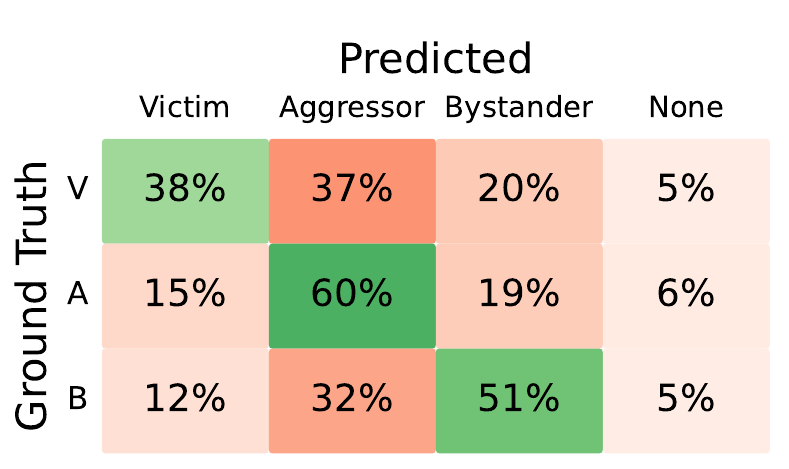}
  \end{minipage}

  \caption{\textbf{Error analysis of action and role association averaged across models.}
  We analyze model failures on Basic questions by decomposing predictions across \textit{(left) action vs. role correctness}, \textit{(center) aggressor vs. victim correctness}, and \textit{(right) role identification}. MLLMs struggle not only to recognize aggressive actions, but also to assign the correct roles to the persons involved.}
  \label{fig:error_analysis}
\end{figure}

\noindent\textbf{\textit{Models often identify participants without recognizing the correct aggressive action.}}
To understand whether errors arise from recognizing the aggressive behavior or identifying the relevant participants, we analyze videos that include both a Basic primary action question and at least one aggressor or victim identification question. Figure~\ref{fig:error_analysis} (left) reports results averaged across models. Models identify the correct aggressor or victim while failing to recognize the correct action in 26\% of cases, whereas they recognize the correct action while failing to identify the relevant role in only 17\% of cases. Another 16\% of cases are correct on both, while 41\% are incorrect on both. This pattern indicates that models are more likely to localize socially salient participants than to distinguish the fine-grained aggressive action being performed. Thus, even when participant grounding partially succeeds, current MLLMs often still lack reliable aggression understanding.

\noindent\textbf{\textit{Models are biased toward assigning the aggressor role.}}
Figures~\ref{fig:error_analysis} (center, right) show a consistent role-grounding failure: models tend to over-assign the aggressor role when identifying participants in aggressive interactions. In Figure~\ref{fig:error_analysis} (center), for videos containing both Basic victim and aggressor identification questions, only 22\% of cases correctly identify the aggressor and the victim simultaneously, while 31\% fail on both roles. Among partial successes, models more often identify the aggressor while missing the victim (28\%) than the other way (19\%), suggesting that victim grounding is unreliable. Figure~\ref{fig:error_analysis} (right) shows the same bias in role classification: many ground-truth victims (37\%) and bystanders (32\%) are predicted as aggressors. These errors suggest that models often treat visual involvement, proximity, or saliency as evidence of aggression, rather than reasoning about who initiates the action and who is targeted.

\noindent\textbf{\textit{Distractor analysis reveals different failure modes across question complexity.}}
Distractor analysis shows that models are especially vulnerable when incorrect options preserve visually plausible participants or partially match the interaction structure. For \textit{Basic} questions, role-reversal distractors are consistently more challenging than cross-video distractors for most models, indicating that models often recognize relevant people but confuse their roles. For \textit{Compound} questions, role-reversal and wrong-action distractors dominate, showing that models struggle to jointly bind the correct action with the correct aggressor or victim. In Detailed questions, frequency-inverted distractors dominate showing that models may rely on answer-choice priors rather than visual grounding, selecting options with familiar or frequently co-occurring action-role-person components instead of recovering the full aggressor-action-victim structure.

Model-level patterns reveal distinct vulnerabilities across question tiers. In \textit{Basic} questions, Ovis2.5-9B is vulnerable to role-reversal distractors, while LLaVA-Video-7B spreads errors across cross-video, wrong-role, and bystander-substitution options, suggesting weaker visual grounding. InternVL3.5-8B-DoT, ViCLIP, and X-CLIP also show high bystander-substitution errors, suggesting over-attention to visible non-participants. In \textit{Compound} questions, ViCLIP, X-CLIP, and LLaVA-Video-7B show more wrong-aggressor and wrong-victim errors than most models, indicating weaker participant grounding. In \textit{Detailed} questions, frequency-inverted distractors dominate for most models, while GPT-5.1 shows more errors involving role reversals.

\noindent\textbf{\textit{Models over-predict aggression even in non-aggressive clips.}}
We conduct a free-form social appropriateness evaluation on four relatively strong models~\cite{ovis2_5, internvl2_5, internvl3, qwen3vl}. Each model is asked to characterize the social appropriateness of the action shown in a video. Models are generally able to identify aggressive or socially inappropriate behavior in aggressive clips, with most models reaching 78.2\%-89.8\% accuracy. However, on non-aggressive clips, they frequently still attribute inappropriate or aggressive behavior to the videos, with most models near or below 60\% accuracy. This indicates that current models are sensitive to aggressive cues but also exhibit a strong aggression prior, over-predicting aggression even when no aggressive behavior is present. InternVL2.5-8B is more balanced across aggressive and non-aggressive clips, but both scores are close to 60\%; given the 50\% binary chance level, this suggests that it does not reliably solve either side of the task. 

\noindent\textbf{\textit{Scene context adds another layer of grounding difficulty.}}
We further evaluate whether models can associate aggressive actions with their surrounding scene context using a subset of videos with clearly identifiable environments. Models are asked to identify both the action and the environment in which it occurs. Adding this scene-grounding requirement makes the task harder for most models: 12 evaluated models show lower accuracy on action-environment association than on primary action recognition, with their average accuracy dropping from 38.56\% to 32.34\%. This indicates that models struggle not only to recognize aggressive actions, but also to ground them in the correct environmental context. Interestingly, InternVL3.5-8B and its reasoning variants are relatively robust, showing little degradation. Overall, however, the drop for most models suggests that reliable action-environment association requires joint reasoning over local behavior and broader scene context. 

Detailed results and analysis charts are provided in the Supplementary.

\subsection{Role-graph prompting}

\begin{figure}[t]
\centering

\begin{minipage}[t]{0.49\textwidth}
    \centering
    \vspace{0pt}
    \includegraphics[width=\linewidth]{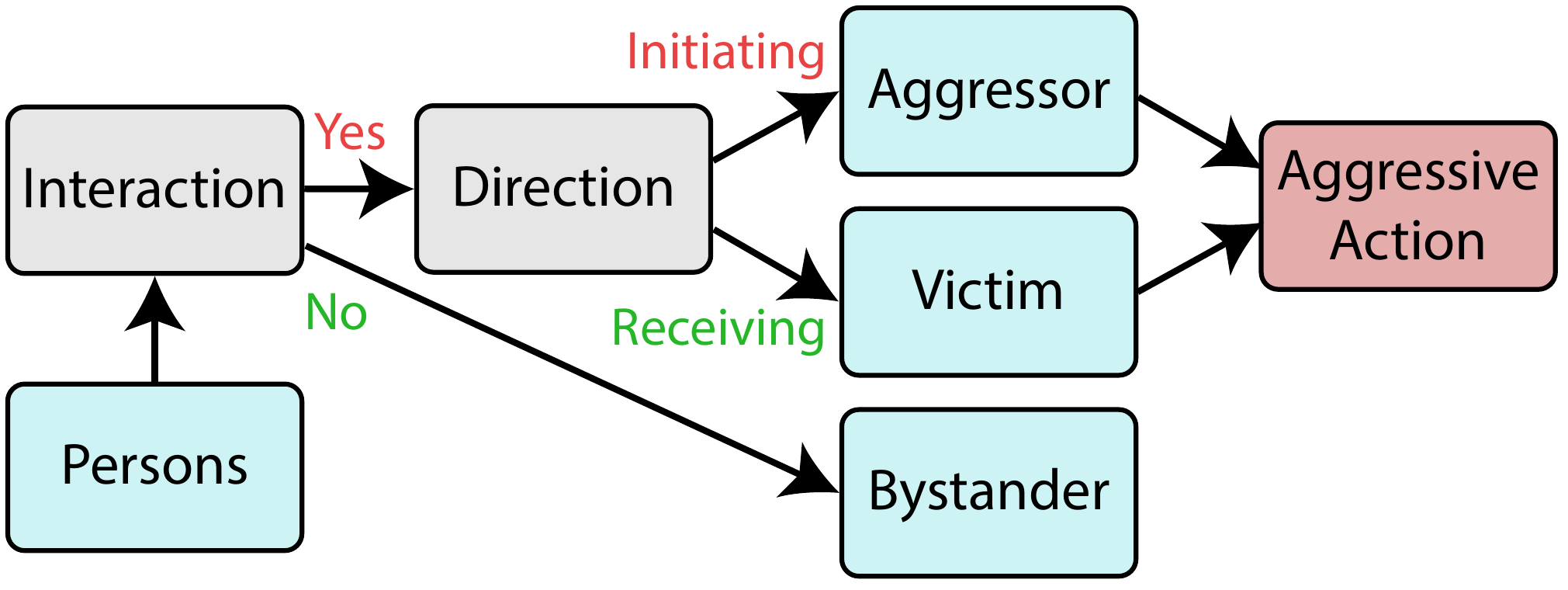}
    \captionof{figure}{\textbf{Role-graph prompting structure.}
    The graph decomposes aggressive-scene understanding into structured role-action reasoning, linking interaction, direction to participant roles and aggressive actions.}
    \label{fig:graph_structure}
\end{minipage}
\hfill
\begin{minipage}[t]{0.49\textwidth}
    \centering
    \vspace{0pt}
    \small
    \captionof{table}{\textbf{Effect of role-graph prompting.}
    Comparison of baseline overall accuracy with role-graph prompting (RG). Results suggest that explicit role-aware prompting can improve aggressive action and participant association, although gains vary across models.
    }
    \label{tab:role_graph_prompt}
    \resizebox{\linewidth}{!}{
    \begin{tabular}{lccc}
    \toprule
    \textbf{Model} & \textbf{Base} \textuparrow & \textbf{RG} \textuparrow  & \textbf{$\Delta$} \\
    \midrule
    InternVL2.5-8B \cite{internvl2_5}       & 45.77 & 48.15 & +2.38 \\
    Ovis2.5-9B \cite{ovis2_5}                & 43.16 & 50.28 & +7.12 \\
    Qwen3-VL-8B-Instruct \cite{qwen3vl}     & 46.96 & 50.65 & +3.69 \\
    Qwen2.5-VL-7B-Instruct \cite{qwen-vl2_5}  & 37.30 & 38.31 & +1.01 \\
    Qwen2.5-VL-72B-Instruct \cite{qwen-vl2_5} & 40.53 & 43.73 & +3.20 \\
    GPT-5.1 \cite{gpt_51}                   & 41.30 & 40.49 & -0.81 \\
    \bottomrule
    \end{tabular}
    }
\end{minipage}

\end{figure}

We observe that models often over-assign the aggressor role, confusing victims or bystanders with active participants. To encourage more explicit role-action reasoning, we explore a role-graph prompting strategy as a diagnostic structured-reasoning intervention. As shown in Figure~\ref{fig:graph_structure}, the graph represents aggressive-scene understanding through interaction direction, participant roles, and aggressive actions. We prepend this graph-based instruction template to each multiple-choice VQA query. The linguistic prompt serves as an interface that directs the model's attention toward interaction direction and participant roles, without altering the underlying visual evidence or evaluation protocol.

Table~\ref{tab:role_graph_prompt} compares baseline accuracy with role-graph prompting (RG) across six models. Role-graph prompting improves performance on five of the six models, with an average accuracy increase from 42.50\% to 45.26\%. Ovis2.5-9B shows the largest improvement, increasing from 43.16\% to 50.28\%. Smaller improvements are observed for InternVL2.5-8B and both Qwen2.5-VL models, while GPT-5.1 shows a slight decrease. These results suggest that explicitly encouraging models to consider participant roles and interaction direction can benefit aggressive-scene understanding, although the effectiveness varies across models. We view this as an exploratory approach to structured role-aware prompting rather than a comprehensive solution to compositional role-action reasoning. The detailed prompt and additional results are provided in the Supplementary.

% Add in preamble if not already included
% \usepackage{caption}

\section{Conclusion}
% We presented A$^3$Bench, a benchmark designed to evaluate whether MLLMs can perform aggressive action and association: recognizing not just that aggression occurs, but who initiates it, who is targeted, and what action connects them. Evaluation of 15 models reveals that current systems plateau below 50\% accuracy and degrade sharply as questions move from single-attribute recognition to full aggressor-action-victim reasoning. A consistent aggressor bias, in which models mislabel victims and bystanders as aggressors, persists across architectures and is not resolved by model scale or reasoning-style prompting. Our role-graph prompting strategy, which decomposes scene understanding into a structured causal chain of interaction detection, direction assignment, and role derivation, improves accuracy by up to 7\%, demonstrating that making the required reasoning can partially compensate this. We intend A$^3$Bench as a diagnostic framework for measuring aggressive action and association.

We propose A$^3$Bench, a benchmark for evaluating whether MLLMs can reason about aggressive actions and participant associations in videos. Unlike prior evaluations focused on aggression detection, A$^3$Bench measures whether models can identify who initiates an action, who is affected, and how participants interact. Our evaluation across 17 models shows that current systems struggle with this task, with performance declining as questions require more compositional reasoning over actions and roles. We further observe an aggressor bias, where victims and bystanders are frequently misclassified as aggressors, suggesting that models fail to reliably capture interaction structure. Finally, we explore a role-graph prompting strategy that encourages structured reasoning over participant interactions and improves performance on most evaluated models. We hope A$^3$Bench serves as a diagnostic benchmark for studying fine-grained role-aware video reasoning.

\begin{ack}
This research received no external funding and the authors declare no competing interests. We gratefully acknowledge Adams Ubini (University of Maryland, Baltimore County), as well as Tarina Sadek and Godfred Obeng (University of Central Florida), for their valuable support and contributions to this work.
\end{ack}

{\small
\bibliographystyle{plain}
\bibliography{egbib}
}

%%%%%%%%%%%%%%%%%%%%%%%%%%%%%%%%%%%%%%%%%%%%%%%%%%%%%%%%%%%%

\appendix

\FloatBarrier
\newpage
\section{Supplementary material}
The supplementary material expands on the main paper with limitations and broader impact discussion, prompt and question-format examples, additional analysis, figures and performance tables, and the full role-graph prompt.

\subsection{Limitations and Broader Impact}
A$^3$Bench focuses specifically on visible physical aggression in short video clips and does not cover broader forms of harmful behavior such as verbal abuse, long-term conflict, or subtle social interactions. Furthermore, A$^3$Bench is designed to evaluate compositional reasoning over aggressive interactions rather than spatial or temporal action detection, and therefore does not provide participant bounding boxes or temporal tracks. Extending the benchmark with such annotations represents an important direction for future work. While A$^3$Bench includes diverse videos, question types, and model families under a controlled evaluation protocol, cross-dataset generalization remains unexplored. Evaluating role-action reasoning across additional datasets and domains is an important direction for future research. This work aims to advance safer and more reliable video understanding systems for applications such as public safety, content moderation, and assistive monitoring. At the same time, aggressive-scene understanding systems may inherit societal biases or produce harmful misclassifications, particularly in high-stakes settings involving surveillance or security. We therefore position A$^3$Bench as a diagnostic evaluation framework rather than a deployment-ready solution.

\subsection{Prompt and Multiple-choice question example}
\begin{figure}[h!]
  \centering
  \includegraphics[width=\linewidth]{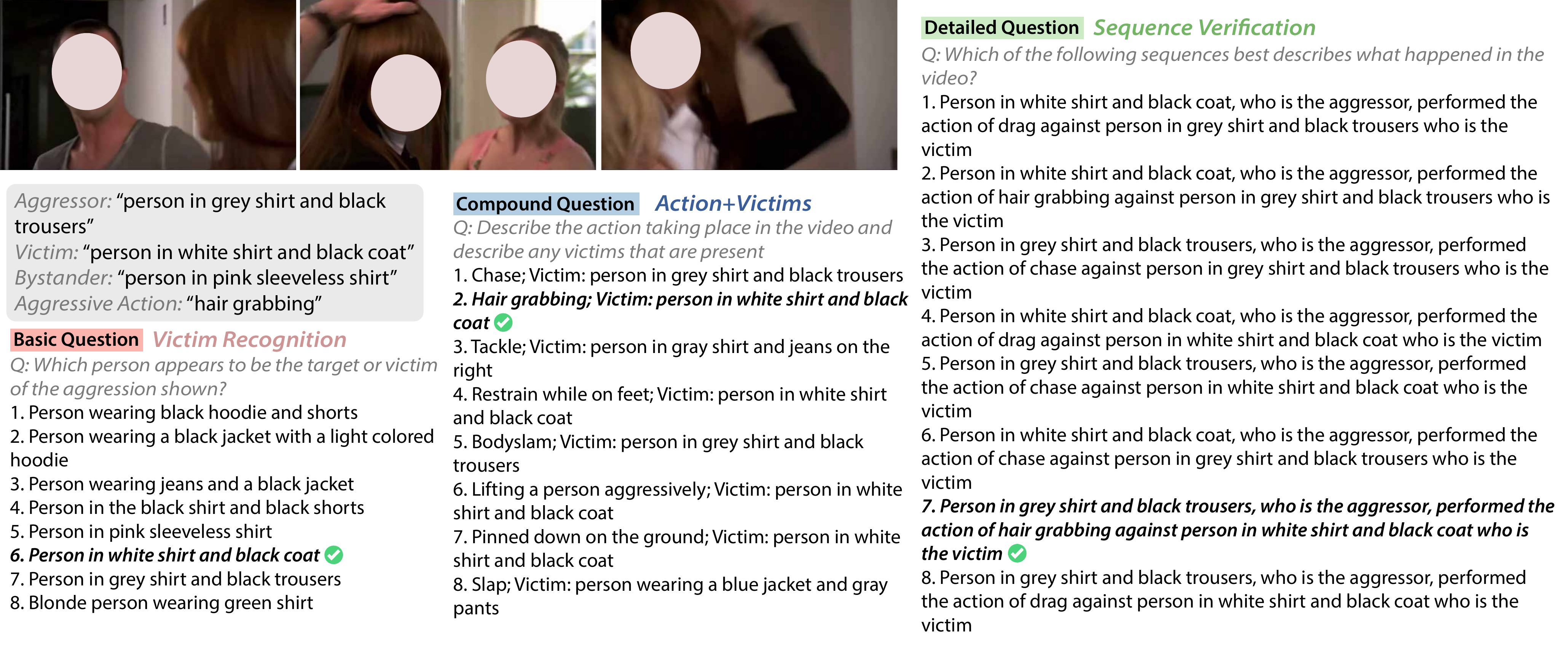}
  \caption{\textbf{Example A$^3$Bench questions across categories and list of choices.}
We show representative structured annotations, and multiple-choice questions generated from a video clip. The left grey panel illustrates the annotated scene information, including aggressor, victim, bystander, and aggressive action descriptions. The bottom-left example shows a \textit{Basic} question that asks for a single attribute, such as identifying the victim. The middle panel shows a \textit{Compound} question that requires binding two attributes, such as the action and victim. The right panel shows a \textit{Detailed} sequence-verification question that requires selecting the full structured event description, including the aggressor, action, victim, and role directionality. Correct answers are highlighted along with green check mark.}
  \label{fig:question_example_detail}
\end{figure}

The prompt given to the model follows the format below, where the question and choice examples are as detailed in Figure~\ref{fig:question_example_detail}.

{\small\ttfamily\raggedright
Watch this video carefully and answer the following multiple-choice question.\\
Select ONLY the number (1, 2, 3, etc.) of the correct answer.\\
Question: \textless question\textgreater\\
Options: \textless choices\textgreater\\
Answer with ONLY the option number (e.g., `1' or `2').
}

\clearpage

\subsection{Additional Analysis}
\label{sec:additional_baselines}

\noindent\textbf{Effect of temporal sampling.}
To assess the importance of temporal information for aggressive action and association, we evaluate Qwen3-VL-8B-Instruct using different numbers of uniformly sampled video frames. Table~\ref{tab:temporal_sampling} reports performance using a single middle frame and 4, 8, and 16 frames, with 8 frames corresponding to our standard evaluation setting. Increasing temporal coverage from a single frame to 16 frames improves overall accuracy from 40.86\% to 51.34\%, with improvements across Basic, Compound, and Detailed questions. While performance is not strictly monotonic across all sampling configurations, these results suggest that temporal information is important for recognizing aggressive actions and associating them with the correct participants, beyond the static appearance and scene cues available from individual frames.

\begin{table}[t]
    \small
    \centering
    \caption{\textbf{Effect of temporal sampling.} Accuracy (\%) of Qwen3-VL-8B-Instruct \cite{qwen3vl} using different numbers of video frames.}
    \label{tab:temporal_sampling}
    \begin{tabular}{lcccc}
        \toprule
        \textbf{Setting} & \textbf{Basic} & \textbf{Compound} & \textbf{Detailed} & \textbf{Overall} \\
        \midrule
        1 (center frame) & 43.94 & 40.55 & 32.25 & 40.86 \\
        4 frames         & 51.98 & 44.98 & 39.72 & 47.45 \\
        8 frames         & 51.07 & 46.65 & 35.19 & 46.96 \\
        16 frames        & 56.77 & 48.56 & 41.62 & 51.34 \\
        \bottomrule
    \end{tabular}
\end{table}

\noindent\textbf{No-video baseline.}
To examine whether models can exploit language or answer-choice priors without visual evidence, we evaluate three MLLMs using only the question and answer options, without providing video frames. As shown in Table~\ref{tab:no_video_baseline}, overall accuracy remains low across all three models, ranging from 19.21\% to 25.07\%. These results indicate that language and answer-choice priors alone are insufficient for reliable performance on A$^3$Bench, further highlighting the importance of visual grounding.

\begin{table}[t]
    \small
    \centering
    \caption{\textbf{No-video baseline.} Accuracy (\%) when models receive only the question and answer choices without video input.}
    \label{tab:no_video_baseline}
    \begin{tabular}{lcccc}
        \toprule
        \textbf{Model} & \textbf{Basic} & \textbf{Compound} & \textbf{Detailed} & \textbf{Overall} \\
        \midrule
        Qwen3-VL-8B-Instruct \cite{qwen3vl}   & 22.66 & 32.45 & 14.29 & 25.07 \\
        LLaVA-Video-7B \cite{llavavideo}      & 17.23 & 19.48 & 24.83 & 19.21 \\
        VideoLLaMA3-7B \cite{videollama3}     & 15.04 & 23.94 & 33.35 & 20.85 \\
        \bottomrule
    \end{tabular}
\end{table}

\noindent\textbf{Performance across participant sex.} To examine potential demographic differences in participant grounding, we evaluate Basic aggressor identification, victim identification, and role identification questions separately according to the sex of the target participant. Averaged equally across all 17 model configurations, accuracy is nearly identical for male and female participants in aggressor identification (48.48\% vs. 47.78\%) and role identification (49.08\% vs. 51.14\%). However, female participants achieve substantially higher accuracy in victim identification (51.33\% vs. 40.13\%), a gap of 11.20\%. This suggests that demographic performance differences are most pronounced in victim identification. Importantly, participant sex is used only for this diagnostic analysis and is not included in the benchmark's role annotations, questions, or answer choices.

\noindent\textbf{Aggressive scenes for fine-grained understanding.}
Table~\ref{tab:nonaggressive_aggressive_results} compares model performance on questions associated with aggressive and non-aggressive clips. Most MLLMs achieve higher accuracy on non-aggressive clips, with pooled accuracy of 51.02\% compared to 37.74\% on aggressive clips. This suggests that models struggle more with fine-grained understanding of aggressive interactions, where identifying actions and correctly associating them with participant roles is challenging. Interestingly, this contrasts with the social appropriateness evaluation in Table~\ref{tab:social_appropriateness}, where models generally perform better on aggressive clips. Together, these findings highlight a distinction between recognizing broad aggressive behavior and accurately reasoning about the actions and participant relationships within aggressive scenes.

\begin{table}[h]
\small
\centering
\caption{\textbf{Performance on non-aggressive and aggressive clips.} Overall accuracy (\%) is reported separately for questions associated with non-aggressive and aggressive video clips. The \textbf{higher score} for each model is highlighted.}
\label{tab:nonaggressive_aggressive_results}
\resizebox{0.65\textwidth}{!}{
\begin{tabular}{l | c c}
\toprule
\textbf{Model}
& \textbf{Non-aggressive} \textuparrow
& \textbf{Aggressive} \textuparrow \\
\midrule
ViCLIP \cite{internvid} & 10.41 & \textbf{24.39} \\
X-CLIP \cite{xclip} & 18.70 & \textbf{20.18} \\
gemma-4-26B-A4B-it \cite{gemma4} & \textbf{73.33} & 41.41 \\
InternVideo2.5-8B \cite{internvideo2_5} & \textbf{54.96} & 40.08 \\
InternVL2.5-8B \cite{internvl2_5} & \textbf{60.49} & 45.14 \\
InternVL3-9B \cite{internvl3} & \textbf{57.07} & 41.93 \\
InternVL3.5-8B \cite{internvl3_5} & \textbf{53.66} & 40.52 \\
LLaVA-Video-7B-Qwen2 \cite{llavavideo} & \textbf{18.54} & 18.49 \\
Ovis2.5-9B \cite{ovis2_5} & \textbf{74.47} & 41.82 \\
Qwen2.5-VL-7B-Instruct \cite{qwen-vl2_5} & \textbf{57.24} & 36.45 \\
Qwen3-VL-8B-Instruct \cite{qwen3vl} & \textbf{57.56} & 46.51 \\
VideoLLaMA3-7B \cite{videollama3} & \textbf{49.43} & 37.74 \\
Qwen2.5-VL-72B-Instruct \cite{qwen-vl2_5} & \textbf{61.79} & 39.62 \\
GPT 5.1 \cite{gpt_51} & \textbf{59.19} & 40.54 \\
\midrule
Qwen3-VL-8B-Thinking \cite{qwen3vl} & \textbf{57.07} & 38.62 \\
InternVL3.5-8B CoT prompt & \textbf{48.29} & 43.76 \\
InternVL3.5-8B DoT \cite{navigatinghallucinations} & \textbf{55.12} & 44.33 \\
\bottomrule
\end{tabular}
}
\end{table}

\subsection{Additional Data and Figures}
We provide additional graphics and detailed numbers for the results and analysis in the main paper.
\begin{figure}[h!]
    \centering
    \includegraphics[width=0.8\linewidth]{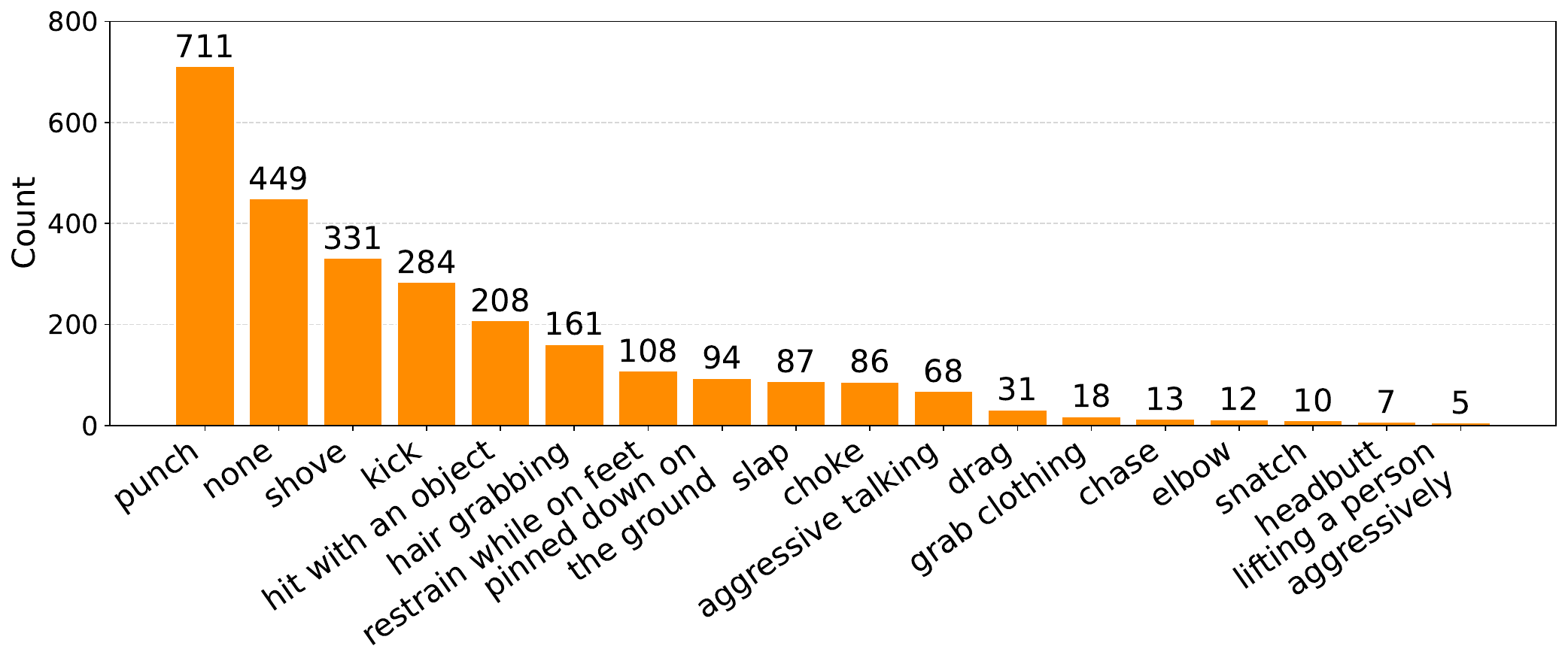}
    \caption{\textbf{Action distribution.} Number of video clips for each annotated aggressive action category in the dataset. This follows a long-tail distribution.}
\end{figure}

\begin{figure}[t!]
  \centering
    \centering
    \includegraphics[width=0.9\linewidth]{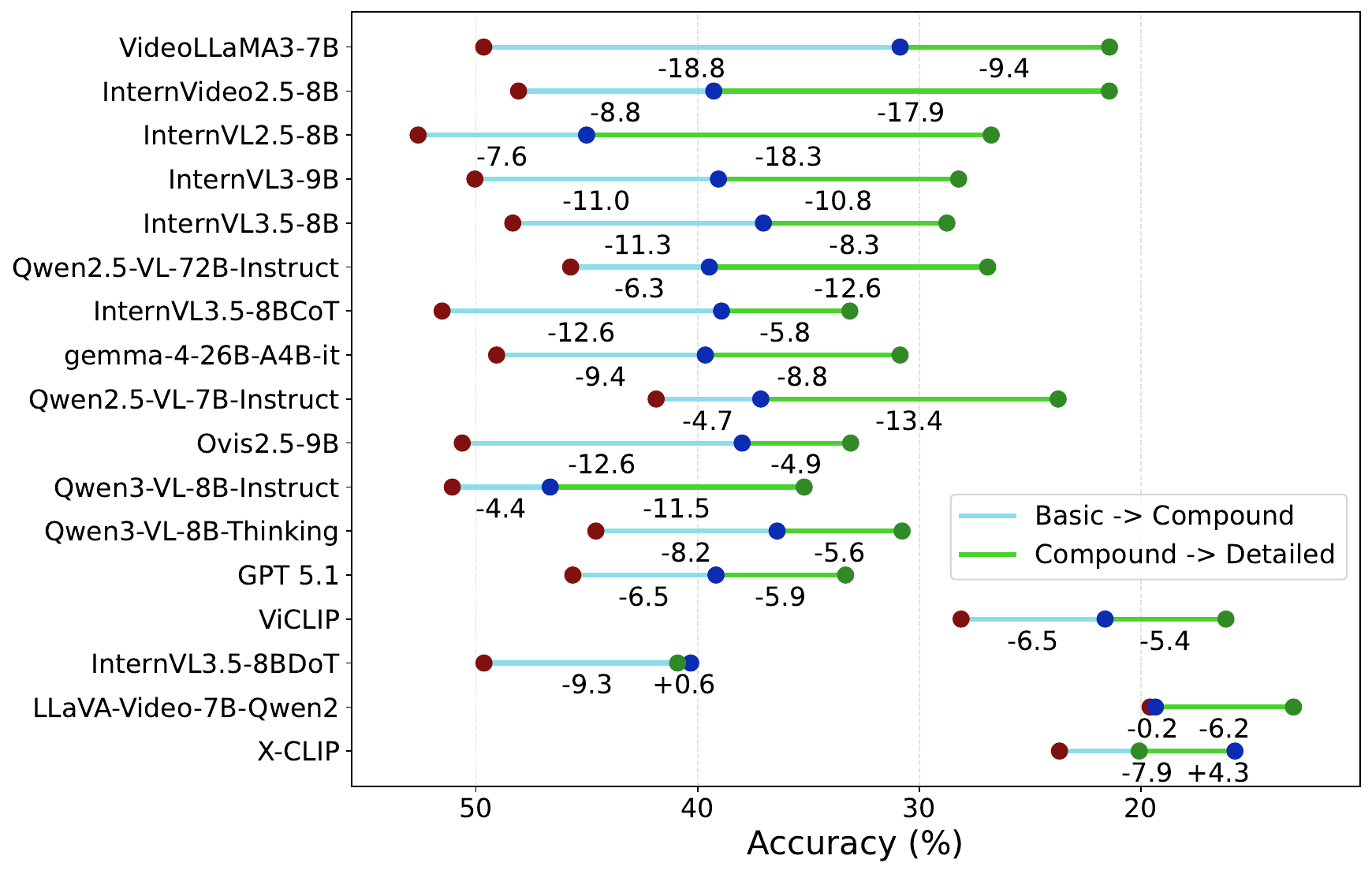}
  \caption{\textbf{Model-wise performance drop.} Comparison of score changes across models for higher level of question complexity.
}
  \label{fig:scores_drop_dumbbell_plot.pdf}
\end{figure}

\begin{table*}[h]
\small
\centering
\caption{\textbf{Question-type-wise performance on A$^3$Bench.} Reported metric is accuracy (\%) for each question subcategory and Overall Accuracy. \textbf{Best} and \underline{second-best} performance are highlighted.}
\label{tab:a3bench_questionwise_results}
\resizebox{\textwidth}{!}{
\begin{tabular}{l | c c c c| c c c| c c| c}
\toprule
\textbf{Model}
& \textbf{Primary}
& \textbf{Role}
& \textbf{Aggressor}
& \textbf{Victim}
& \textbf{Action+}
& \textbf{Action+}
& \textbf{Aggressor+}
& \textbf{Aggressor+Action}
& \textbf{Sequence}
& \textbf{Overall} \\
& \textbf{Action}
& \textbf{ID}
& \textbf{ID}
& \textbf{Recog.}
& \textbf{Aggressors}
& \textbf{Victims}
& \textbf{Victim}
& \textbf{+Victim}
& \textbf{Verification}
& \textbf{Accuracy} \\
\midrule
ViCLIP \cite{internvid} & 25.08 & 24.78 & 33.86 & 30.92 & 22.99 & 25.61 & 11.65 & 18.40 & 13.67 & 23.82 \\
X-CLIP \cite{xclip} & 22.97 & 25.63 & 23.25 & 21.99 & 16.16 & 18.46 & 9.98 & 20.13 & 20.00 & 20.12 \\
gemma-4-26B-A4B-it \cite{gemma4} & \textbf{48.92} & 48.81 & 53.10 & 45.50 & 37.92 & 36.80 & 48.13 & 30.00 & 31.81 & 42.72 \\
InternVideo2.5-8B \cite{internvideo2_5} & 43.70 & 54.68 & 49.67 & 41.77 & 37.39 & 38.98 & 43.64 & 21.32 & 21.72 & 40.70 \\
InternVL2.5-8B \cite{internvl2_5} & 46.00 & \underline{56.02} & 55.89 & 52.13 & \underline{42.15} & \textbf{45.59} & \textbf{49.29} & 25.99 & 27.59 & \underline{45.77} \\
InternVL3-9B \cite{internvl3} & 40.34 & 55.71 & 53.19 & 50.10 & 39.93 & 35.28 & 44.39 & 30.94 & 25.21 & 42.55 \\
InternVL3.5-8B \cite{internvl3_5} & 36.13 & 53.61 & 53.85 & 49.63 & 35.98 & 32.43 & 47.46 & 28.80 & 28.69 & 41.06 \\
LLaVA-Video-7B-Qwen2 \cite{llavavideo} & 19.37 & 23.11 & 20.67 & 13.39 & 16.19 & 15.51 & 32.25 & 13.20 & 13.02 & 18.50 \\
Ovis2.5-9B \cite{ovis2_5} & 39.84 & 54.81 & 55.63 & \underline{52.27} & 35.63 & 34.83 & 48.21 & 33.50 & 32.63 & 43.16 \\
Qwen2.5-VL-7B-Instruct \cite{qwen-vl2_5} & 41.48 & 42.90 & 43.94 & 38.66 & 36.77 & 34.18 & 43.31 & 24.96 & 22.36 & 37.30 \\
Qwen3-VL-8B-Instruct \cite{qwen3vl} & \underline{46.47} & 55.42 & 53.06 & 48.04 & \textbf{49.27} & \underline{43.89} & 46.80 & \underline{34.60} & \underline{35.84} & \textbf{46.96} \\
VideoLLaMA3-7B \cite{videollama3} & 43.34 & 53.87 & 53.98 & 46.52 & 22.63 & 35.05 & 38.57 & 20.96 & 21.91 & 38.22 \\
Qwen2.5-VL-72B-Instruct \cite{qwen-vl2_5} & 39.00 & 50.54 & 50.59 & 41.67 & 39.22 & 35.10 & \underline{48.67} & 26.53 & 27.34 & 40.53 \\
GPT 5.1 \cite{gpt_51} & 44.06 & 51.66 & 45.59 & 38.50 & 39.77 & 37.87 & 40.43 & 33.75 & 32.84 & 41.30 \\
\midrule
Qwen3-VL-8B-Thinking \cite{qwen3vl} & 33.70 & 51.64 & 51.19 & 40.42 & 35.44 & 32.19 & 45.97 & 30.41 & 31.16 & 39.37 \\
InternVL3.5-8B CoT prompt & 37.74 & \textbf{58.53} & \textbf{58.20} & 50.91 & 38.60 & 36.23 & 44.47 & 32.51 & 33.82 & 43.94 \\
InternVL3.5-8B DoT \cite{navigatinghallucinations} & 37.85 & 51.86 & \underline{57.08} & \textbf{53.01} & 40.15 & 36.64 & 47.38 & \textbf{39.77} & \textbf{42.16} & 44.77 \\
\bottomrule
\end{tabular}
}
\end{table*}

\begin{table}[h]
\small
\centering
\caption{\textbf{Action recognition with and without scene grounding.} Accuracy (\%) is reported for Primary Action and Action+Location questions. The \textbf{higher score} for each model is highlighted.}
\label{tab:action_location_results}
\resizebox{0.65\textwidth}{!}{
\begin{tabular}{l | c c}
\toprule
\textbf{Model}
& \textbf{Primary Action} \textuparrow
& \textbf{Action+Location} \textuparrow \\
\midrule
ViCLIP \cite{internvid} & 25.08 & \textbf{26.50}\\
X-CLIP \cite{xclip} & \textbf{22.97} & 15.95\\
gemma-4-26B-A4B-it \cite{gemma4} & \textbf{48.92} & 31.90 \\
InternVideo2.5-8B \cite{internvideo2_5} & \textbf{43.70} & 33.53 \\
InternVL2.5-8B \cite{internvl2_5} & \textbf{46.00} & 38.22 \\
InternVL3-9B \cite{internvl3} & \textbf{40.34} & 35.90 \\
InternVL3.5-8B \cite{internvl3_5} & 36.13 & \textbf{38.99} \\
LLaVA-Video-7B-Qwen2 \cite{llavavideo} & \textbf{19.37} & 17.91 \\
Ovis2.5-9B \cite{ovis2_5} & \textbf{39.84} & 37.70 \\
Qwen2.5-VL-7B-Instruct \cite{qwen-vl2_5} & \textbf{41.48} & 33.02 \\
Qwen3-VL-8B-Instruct \cite{qwen3vl} & 46.47 & \textbf{46.74} \\
VideoLLaMA3-7B \cite{videollama3} & \textbf{43.34} & 35.90 \\
Qwen2.5-VL-72B-Instruct \cite{qwen-vl2_5} & \textbf{39.00} & 35.71 \\
GPT 5.1 \cite{gpt_51} & \textbf{44.06} & 40.91 \\
\midrule
Qwen3-VL-8B-Thinking \cite{qwen3vl} & \textbf{33.70} & 31.39 \\
InternVL3.5-8B CoT prompt & 37.74 & \textbf{38.30} \\
InternVL3.5-8B DoT \cite{navigatinghallucinations} & 37.85 & \textbf{40.70} \\
\bottomrule
\end{tabular}
}
\end{table}

\begin{figure}[h!]
  \centering
  \includegraphics[width=\linewidth]{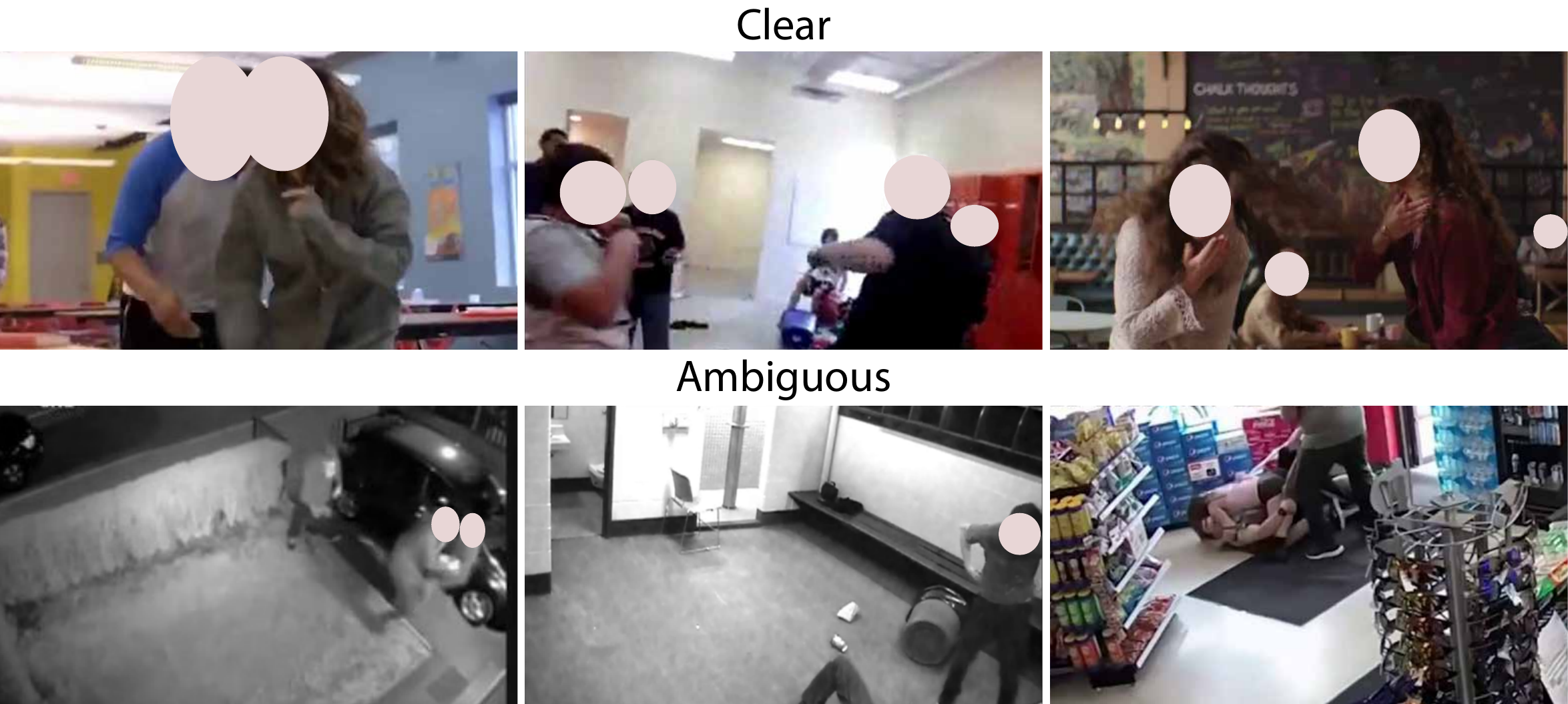}
  \caption{\textbf{Qualitative examples of clear and ambiguous participant grounding.} Top: Clear examples where aggressors and victims are visually distinguishable through appearance, clothing, and interaction cues, enabling reliable role annotations. Bottom: Ambiguous examples where low-quality or black-and-white footage, occlusion, and limited visibility make participant identification difficult. Uncertain role descriptions are left empty rather than guessed, while clips with insufficient visual evidence are excluded from the dataset.}
  \label{fig:annotation_examples}
\end{figure}

\begin{table}[t]
\small
\centering
\caption{\textbf{Social appropriateness binary classification.} Accuracy (\%) is reported separately for aggressive and non-aggressive clips. The \textbf{higher score} for each model is highlighted.}
\label{tab:social_appropriateness}
\resizebox{0.62\textwidth}{!}{
\begin{tabular}{lcc}
\toprule
\textbf{Model}
& \textbf{Aggressive} \textuparrow
& \textbf{Non-aggressive} \textuparrow \\
\midrule
Qwen3-VL-8B \cite{qwen3vl} & \textbf{89.8} & 40.9 \\
Ovis2.5-9B \cite{ovis2_5} & \textbf{78.2} & 60.9 \\
InternVL3-9B \cite{internvl3} & \textbf{78.5} & 55.3 \\
InternVL2.5-8B \cite{internvl2_5} & 61.0 & \textbf{65.5} \\
\bottomrule
\end{tabular}
}
\end{table}

\begin{figure}[h!]
  \centering
  \includegraphics[width=\linewidth]{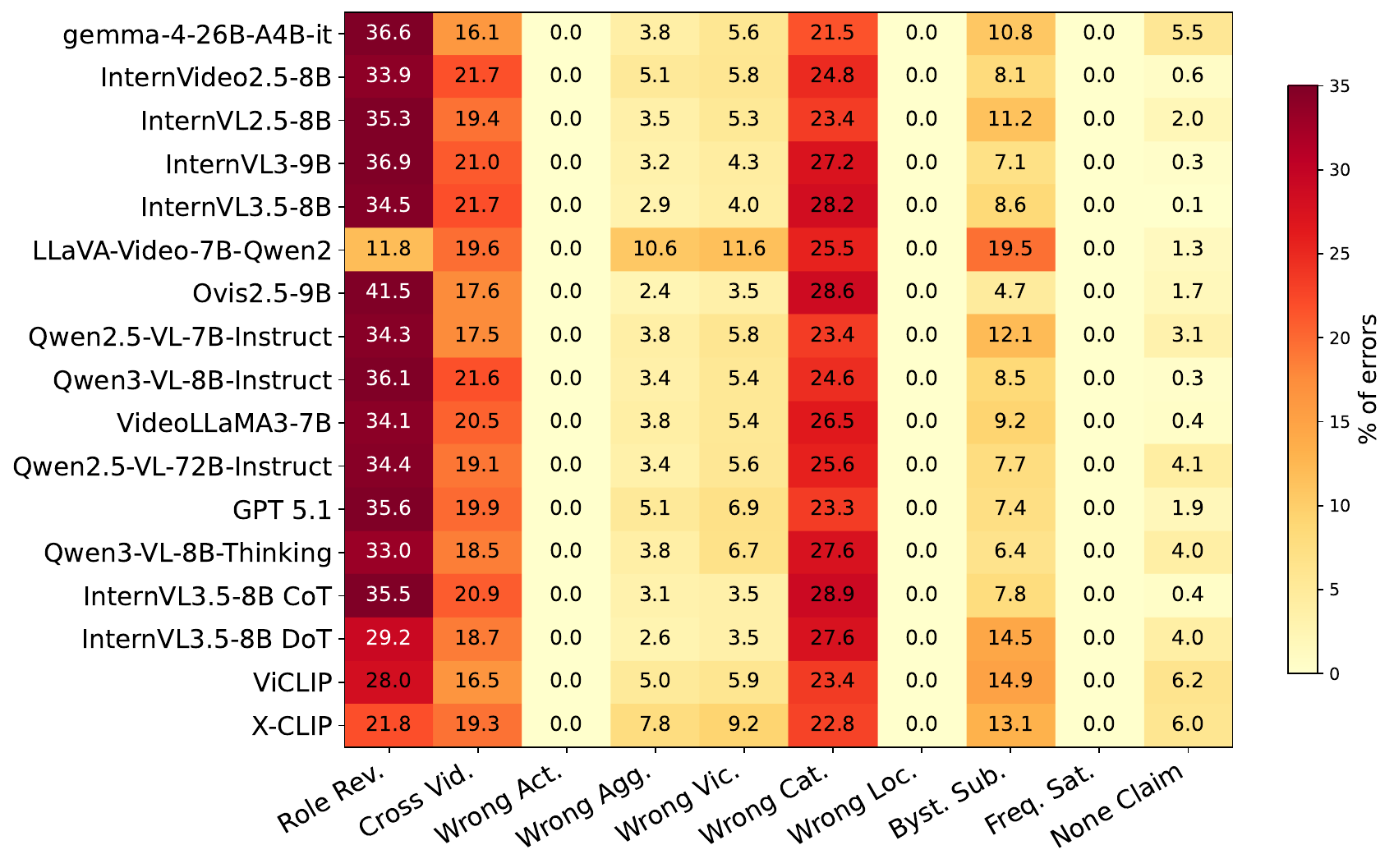}
  \caption{\textbf{Distractor vulnerability heatmap for Basic questions.} Each cell shows the percentage of a model's errors attributed to each distractor type.}
\end{figure}
\begin{figure}[h!]
  \centering
  \includegraphics[width=\linewidth]{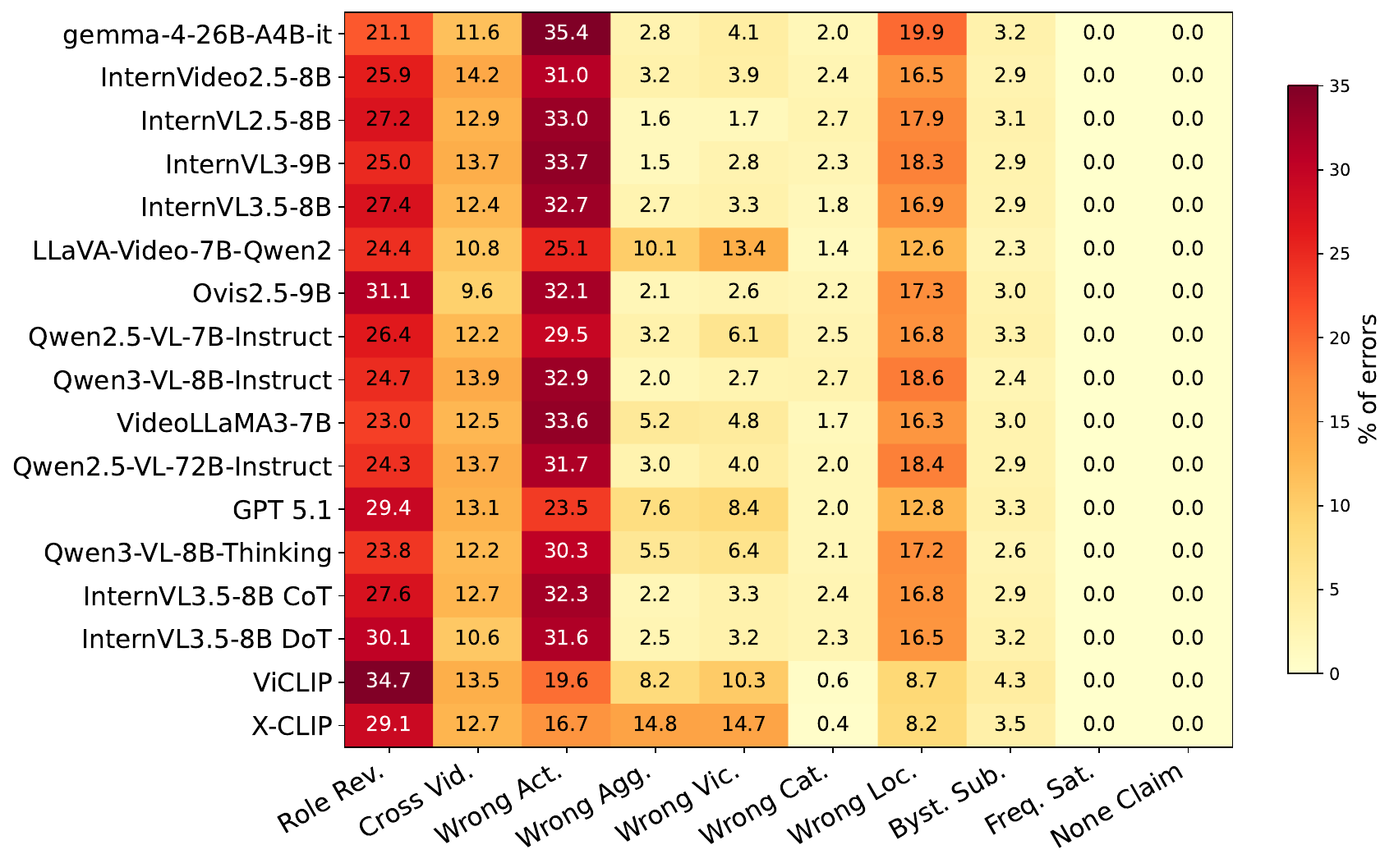}
  \caption{\textbf{Distractor vulnerability heatmap for Compound questions.} Each cell shows the percentage of a model's errors attributed to each distractor type.}
\end{figure}
\begin{figure}[h!]
  \includegraphics[width=\linewidth]{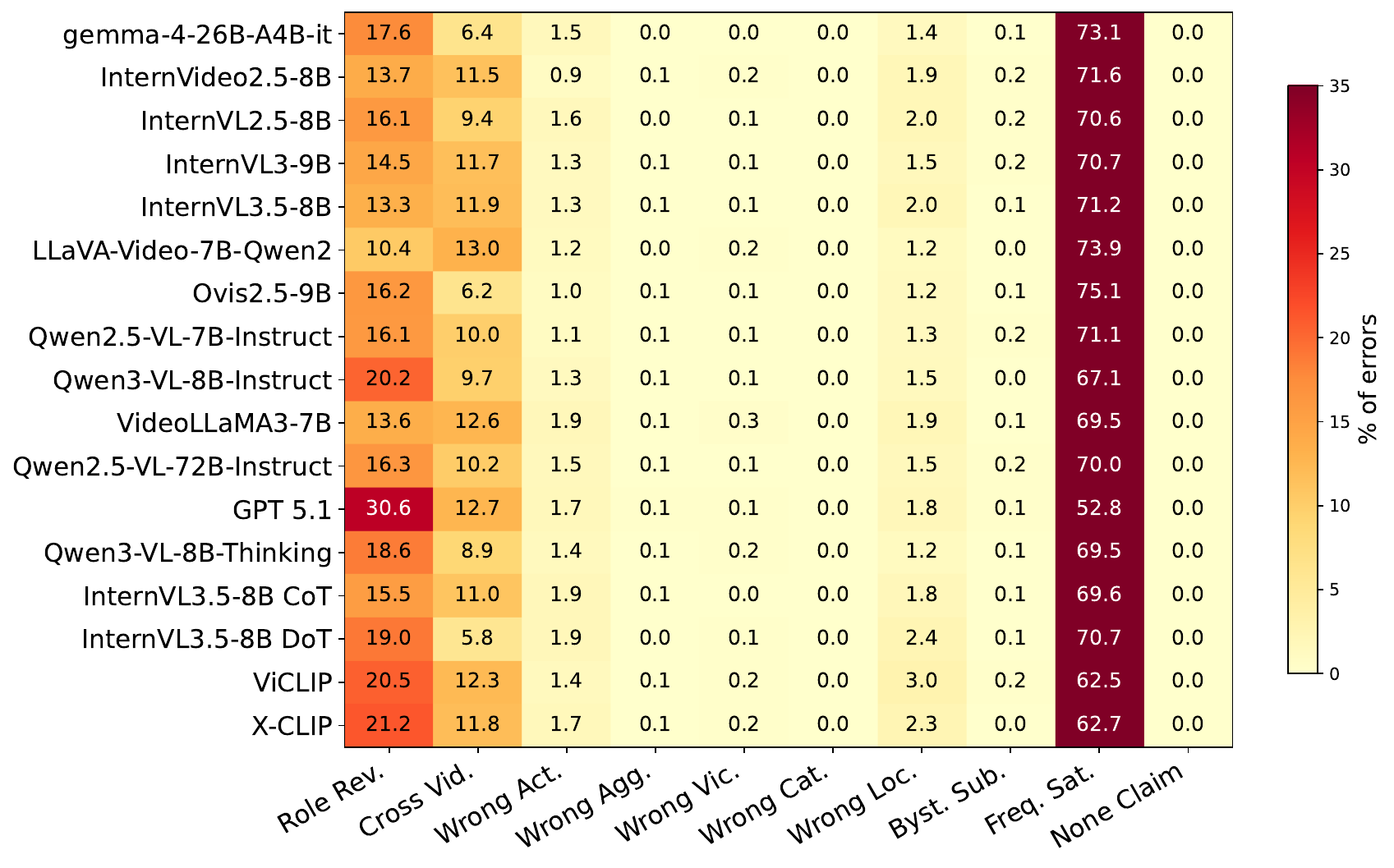}
  \caption{\textbf{Distractor vulnerability heatmap for Detailed questions.} Each cell shows the percentage of a model's errors attributed to each distractor type.}
\end{figure}

\FloatBarrier

\begin{table}[h]
\small
\centering
\caption{\textbf{Role-graph prompting results on A$^3$Bench.} Reported metric is accuracy (\%) for Basic Questions, Compound Questions, Detailed Questions, Overall, and Average. \textbf{Best} and \underline{second-best} performance are highlighted.}
\label{tab:role_graph_results}
\resizebox{\textwidth}{!}{
\begin{tabular}{l | c c c c c}
\toprule
\textbf{Model}
& \textbf{Basic} \textuparrow
& \textbf{Compound} \textuparrow
& \textbf{Detailed} \textuparrow
& \textbf{Overall} \textuparrow
& \textbf{Average} \textuparrow \\
\midrule
InternVL2.5-8B \cite{internvl2_5} & 54.54 & 46.35 & 33.06 & 48.15 & 44.65 \\
Ovis2.5-9B \cite{ovis2_5} & \textbf{55.08} & \underline{46.94} & \textbf{43.83} & \underline{50.28} & \underline{48.62} \\
Qwen3-VL-8B-Instruct \cite{qwen3vl} & \underline{54.81} & \textbf{48.65} & \underline{42.88} & \textbf{50.65} & \textbf{48.78} \\
Qwen2.5-VL-7B-Instruct \cite{qwen-vl2_5} & 37.27 & 41.00 & 34.83 & 38.31 & 37.70 \\
Qwen2.5-VL-72B-Instruct \cite{qwen-vl2_5} & 46.56 & 44.56 & 33.06 & 43.73 & 41.39 \\
GPT-5.1 \cite{gpt_51} & 44.14 & 37.95 & 35.58 & 40.49 & 39.22 \\
\bottomrule
\end{tabular}
}
\end{table}

\subsection{Role-Graph Prompting}

We translate the proposed role graph into a structured natural-language prompt for aggressive-scene understanding. The prompt defines each graph component, specifies the directed dependencies between components, and provides illustrative examples for aggressive scenes, bystander cases, and non-aggressive scenes. During evaluation, this role-graph prompt is prepended to each multiple-choice VQA query, and the model is instructed to output only the selected option number. The exact prompt used in our experiments is provided below.

{\small\ttfamily\raggedright
Your primary task is to act as a causal reasoning engine for understanding aggressive scenes in videos. You will observe the persons present and use a causal graph to identify each person's role and the nature of the aggressive interaction.

\noindent\textbf{\#\#\# Component Definitions}

First, understand the meaning of each component (node) in our causal graph. These are the building blocks we use to analyze any aggressive scene.

\begin{itemize}
    \item \textbf{P (Persons):} All individuals visible in the video. For example, `a man in a black coat', `a woman in a red shirt', `three people in the background'. Each person must be assigned exactly one role.
    \item \textbf{I (Interaction):} Whether a given person is engaged in a direct physical interaction with another person. This is a binary branch point:
    \begin{itemize}
        \item YES — the person is physically engaged with at least one other person, e.g., striking, grabbing, restraining, being struck.
        \item NO — the person has no physical contact with any other person in the scene.
    \end{itemize}
    \item \textbf{D (Direction):} For persons where I = YES, the direction of the aggressive interaction:
    \begin{itemize}
        \item INITIATING — the person is the one delivering the aggressive act, e.g., throwing a punch, grabbing, pushing.
        \item RECEIVING — the person is the one on the receiving end of the aggressive act, e.g., being punched, being grabbed, being pushed.
    \end{itemize}
    \item \textbf{R (Role):} The role assigned to each person, inferred from I and D:
    \begin{itemize}
        \item Aggressor — a person where I = YES and D = INITIATING.
        \item Victim — a person where I = YES and D = RECEIVING.
        \item Bystander — a person where I = NO, not physically involved.
    \end{itemize}
    \item \textbf{A (Aggressive Action):} The specific type of aggressive behavior being performed, inferred from the Aggressor's motions and the Victim's reactions. For example, `punch', `shove', `kick', `choke', `hair grab'. This is determined jointly by the Aggressor and Victim roles.
\end{itemize}

\noindent\textbf{\#\#\# The Causal Graph: How Components are Related}

The components are connected in a Directed Acyclic Graph (DAG). Each arrow ($\rightarrow$) represents a direction of inference — follow the graph in order to assign roles correctly.

The directed edges of the graph are:

$P \rightarrow I$

$I(\text{YES}) \rightarrow D$

$I(\text{NO}) \rightarrow R(\text{Bystander})$

$D(\text{INITIATING}) \rightarrow R(\text{Aggressor})$

$D(\text{RECEIVING}) \rightarrow R(\text{Victim})$

$R(\text{Aggressor}) + R(\text{Victim}) \rightarrow A$

Here are the causal relationships explained in detail:

\begin{itemize}
    \item $P \rightarrow I$ (Does this person interact?): For every person identified in P, ask whether they are physically interacting with another person. This is the first branch point. A person standing and watching is I = NO. A person throwing a punch is I = YES.
    \item $I(\text{YES}) \rightarrow D$ (What is the direction of their interaction?): For persons with I = YES, determine whether they are initiating or receiving the physical contact. Watch carefully — the aggressor moves toward the victim with force; the victim reacts to the aggressor's action.
    \item $I(\text{NO}) \rightarrow R(\text{Bystander})$: Any person with no physical involvement is immediately classified as a Bystander. No further reasoning is needed for them.
    \item $D(\text{INITIATING}) \rightarrow R(\text{Aggressor})$: A person who is initiating the aggressive physical contact is the Aggressor.
    \item $D(\text{RECEIVING}) \rightarrow R(\text{Victim})$: A person who is on the receiving end of the aggressive physical contact is the Victim.
    \item $R(\text{Aggressor}) + R(\text{Victim}) \rightarrow A$ (What is the Aggressive Action?): Only after both the Aggressor and Victim are identified can the Aggressive Action be determined. Observe the Aggressor's body motion, e.g., arm swinging forward, and the Victim's reaction, e.g., head snapping back, together to name the action precisely.
\end{itemize}

\noindent\textbf{\#\#\# Illustrative Examples}

\textbf{Example 1: A clear two-person aggressive scene}

A video shows two people. One person in a black jacket swings their fist at a person in a white shirt, who stumbles backward.

\begin{itemize}
    \item Causal Analysis (internal, do not output):
    \begin{enumerate}
        \item P: Two persons — `person in a black jacket', `person in a white shirt'.
        \item I for `person in a black jacket': YES (physical contact with fist). I for `person in a white shirt': YES (receiving the fist).
        \item D for `person in a black jacket': INITIATING (swinging the fist). D for `person in a white shirt': RECEIVING (stumbling from the impact).
        \item R: `person in a black jacket' $\rightarrow$ Aggressor. `person in a white shirt' $\rightarrow$ Victim.
        \item A: The Aggressor's motion (arm swinging forward, fist connecting) and Victim's reaction (head snapping, stumbling) $\rightarrow$ Aggressive Action = `punch'.
    \end{enumerate}
\end{itemize}

\textbf{Example 2: A scene with a bystander}

A video shows three people. One person in a red hoodie grabs another person in a grey jacket by the collar. A third person in a blue shirt stands nearby watching.

\begin{itemize}
    \item Causal Analysis (internal, do not output):
    \begin{enumerate}
        \item P: Three persons — `person in a red hoodie', `person in a grey jacket', `person in a blue shirt'.
        \item I for `person in a red hoodie': YES. I for `person in a grey jacket': YES. I for `person in a blue shirt': NO.
        \item D for `person in a red hoodie': INITIATING (grabbing). D for `person in a grey jacket': RECEIVING (being grabbed).
        \item R: `person in a red hoodie' $\rightarrow$ Aggressor. `person in a grey jacket' $\rightarrow$ Victim. `person in a blue shirt' $\rightarrow$ Bystander (I = NO, skip D).
        \item A: The Aggressor grabs the Victim's collar and pulls — Aggressive Action = `clothing grab'.
    \end{enumerate}
\end{itemize}

\textbf{Example 3: No aggressive action}

A video shows two people standing and talking to each other without any physical contact.

\begin{itemize}
    \item Causal Analysis (internal, do not output):
    \begin{enumerate}
        \item P: Two persons.
        \item I for both persons: NO (no physical contact with any other person).
        \item Since I = NO for all persons, no one reaches the D step.
        \item R: Both persons $\rightarrow$ Bystander (no aggressor, no victim).
        \item A: No Aggressor and no Victim exist $\rightarrow$ No aggressive action is taking place.
    \end{enumerate}
\end{itemize}

\noindent\textbf{\#\#\# Your Task}

You will now be given a new video. Apply the causal graph above to reason about every person you see.

Reason through these steps internally. Do not output your reasoning.

\begin{itemize}
    \item \textbf{P (Persons):} Identify all individuals in the video. For each person, carefully describe their appearance — focus first on clothing (color and type, e.g., `black coat', `white t-shirt', `blue jeans'), then hair color/style, then any other distinguishing feature. These descriptions will serve as each person's identifier throughout the rest of your reasoning, and must match the appearance-based labels used in the answer options.
    \item \textbf{I (Interaction):} For each person, determine YES or NO — are they physically interacting with another person?
    \item \textbf{D (Direction):} For each person with I = YES, determine INITIATING or RECEIVING.
    \item \textbf{R (Role):} Assign each person a role — Aggressor, Victim, or Bystander — following the graph edges.
    \item \textbf{A (Aggressive Action):} Using the identified Aggressor and Victim, determine the specific aggressive action. If no Aggressor and Victim exist, the answer is that no aggressive action is taking place.
\end{itemize}

From the list of options below, select the answer that most accurately reflects the question asked about the video.
There might also a choice if none of the other options apply.
Review the video and options carefully. Only one choice is correct.

\textbf{Output format:} reply with the choice number only.
}

%%%%%%%%%%%%%%%%%%%%%%%%%%%%%%%%%%%%%%%%%%%%%%%%%%%%%%%%%%%%

\newpage

\section*{NeurIPS Paper Checklist}
\begin{enumerate}

\item {\bf Claims}
    \item[] Question: Do the main claims made in the abstract and introduction accurately reflect the paper's contributions and scope?
    \item[] Answer: \answerYes{}
    \item[] Justification: The abstract and introduction state three contributions: The A$^3$Bench dataset, the diagnostic evaluation of 17 multimodal models, and the role-graph prompting strategy, all of which are supported by experimental results in Section~4.

\item {\bf Limitations}
    \item[] Question: Does the paper discuss the limitations of the work performed by the authors?
    \item[] Answer: \answerYes{}
    \item[] Justification: The supplementary material discuss limitations including the dataset's focus on physical aggression (excluding verbal or online aggression), annotations, and its positioning as a diagnostic benchmark rather than end-to-end solution.

\item {\bf Theory assumptions and proofs}
    \item[] Question: For each theoretical result, does the paper provide the full set of assumptions and a complete (and correct) proof?
    \item[] Answer: \answerNA{}
    \item[] Justification: This paper is an empirical benchmark and evaluation study. It does not include theoretical results, theorems, or proofs.

    \item {\bf Experimental result reproducibility}
    \item[] Question: Does the paper fully disclose all the information needed to reproduce the main experimental results of the paper to the extent that it affects the main claims and/or conclusions of the paper (regardless of whether the code and data are provided or not)?
    \item[] Answer: \answerYes{}
    \item[] Justification: The question generation procedure, distractor construction, and answer-shuffling protocol are described in Section~3.1. Sections~3.2 and 3.3 specify all model names, versions, GPU hardware, and evaluation protocol. All evaluated models use publicly available weights or official APIs. The role-graph prompt template is provided in the supplementary material.

\item {\bf Open access to data and code}
    \item[] Question: Does the paper provide open access to the data and code, with sufficient instructions to faithfully reproduce the main experimental results, as described in supplemental material?
    \item[] Answer: \answerYes{}
    \item[] Justification: The benchmark dataset (annotations, question bank, and evaluation code) has been publicly released on GitHub. Video clips are distributed in Huggingface under a research-use license.

\item {\bf Experimental setting/details}
    \item[] Question: Does the paper specify all the training and test details (e.g., data splits, hyperparameters, how they were chosen, type of optimizer) necessary to understand the results?
    \item[] Answer: \answerYes{}
    \item[] Justification: Section~3.3 reports the models, and evaluation protocol. No training is performed; all models are evaluated zero-shot using their released weights. The role-graph prompting strategy (Section~4.2) is training-free and its full prompt is in the supplementary.

\item {\bf Experiment statistical significance}
    \item[] Question: Does the paper report error bars suitably and correctly defined or other appropriate information about the statistical significance of the experiments?
    \item[] Answer: \answerNo{}
    \item[] Justification: All evaluations are deterministic (greedy decoding, fixed frame sampling, fixed answer-position shuffling), so there is minimal run-to-run variance to report. Each model is evaluated once on the full benchmark of over 15K questions, providing sufficient sample size for stable accuracy estimates.

\item {\bf Experiments compute resources}
    \item[] Question: For each experiment, does the paper provide sufficient information on the computer resources (type of compute workers, memory, time of execution) needed to reproduce the experiments?
    \item[] Answer: \answerYes{}
    \item[] Justification: Section~3.3 specifies the models and compute.

\item {\bf Code of ethics}
    \item[] Question: Does the research conducted in the paper conform, in every respect, with the NeurIPS Code of Ethics \url{https://neurips.cc/public/EthicsGuidelines}?
    \item[] Answer: \answerYes{}
    \item[] Justification: The research conforms to the NeurIPS Code of Ethics. The dataset contains videos of aggressive behavior collected for research purposes. Person descriptions use clothing and coarse visual appearance only, avoiding names, faces, demographic attributes, or other personally identifying information. The benchmark is designed to evaluate and improve safety-critical AI capabilities.

\item {\bf Broader impacts}
    \item[] Question: Does the paper discuss both potential positive societal impacts and negative societal impacts of the work performed?
    \item[] Answer: \answerYes{}
    \item[] Justification: The paper discusses positive impacts including improved safety monitoring and public safety applications. It also acknowledges potential negative impacts: the aggressor-bias finding highlights that deploying current models could systematically misidentify victims as aggressors, with serious consequences in safety-critical settings.

\item {\bf Safeguards}
    \item[] Question: Does the paper describe safeguards that have been put in place for responsible release of data or models that have a high risk for misuse (e.g., pre-trained language models, image generators, or scraped datasets)?
    \item[] Answer: \answerYes{}
    \item[] Justification: The dataset contains sensitive content (videos of aggressive interactions) from public sources. Person descriptions deliberately avoid personally identifying information such as names, faces, or demographic attributes. The dataset is released under a research-use license with terms prohibiting re-identification of individuals and restricting use to academic research.

\item {\bf Licenses for existing assets}
    \item[] Question: Are the creators or original owners of assets (e.g., code, data, models), used in the paper, properly credited and are the license and terms of use explicitly mentioned and properly respected?
    \item[] Answer: \answerYes{}
    \item[] Justification: We collected videos from social-media and trimmed them for research evaluation purposes, which fall under fair usage policy. All source datasets (UCF-Crime, RWF-2000) and evaluated models are cited with their original publications. We are the original owners of the annotations. All models are used via their publicly released weights and official implementations.

\item {\bf New assets}
    \item[] Question: Are new assets introduced in the paper well documented and is the documentation provided alongside the assets?
    \item[] Answer: \answerYes{}
    \item[] Justification: A$^3$Bench is a new benchmark dataset. Section~3 documents the video curation, annotation procedure, quality assurance process, question generation, and distractor design in detail. The supplementary material includes the full annotation schema, question type definitions, and dataset statistics. The dataset is released with appropriate documentation.

\item {\bf Crowdsourcing and research with human subjects}
    \item[] Question: For crowdsourcing experiments and research with human subjects, does the paper include the full text of instructions given to participants and screenshots, if applicable, as well as details about compensation (if any)?
    \item[] Answer: \answerYes{}
    \item[] Justification: The dataset annotations were produced by a group of annotators following structured guidelines. The annotation instructions, including role definitions, labeling criteria, and quality assurance procedures, are described in Section~3.1. We also conducted an anonymous human evaluation involving 34 participants, with the evaluation procedure and results described in detail in Section~4.

\item {\bf Institutional review board (IRB) approvals or equivalent for research with human subjects}
    \item[] Question: Does the paper describe potential risks incurred by study participants, whether such risks were disclosed to the subjects, and whether Institutional Review Board (IRB) approvals (or an equivalent approval/review based on the requirements of your country or institution) were obtained?
    \item[] Answer: \answerNA{}
    \item[] Justification: The annotators labeled publicly available video content and were not themselves study participants or subjects of research. The annotation task involves labeling observable actions and roles in video clips, not collecting personal data from the annotators. Separately, we conducted a human evaluation involving 34 participants through an anonymous webpage, where participants independently answered video-based multiple-choice questions. No personally identifying information was collected during the evaluation.

\item {\bf Declaration of LLM usage}
    \item[] Question: Does the paper describe the usage of LLMs if it is an important, original, or non-standard component of the core methods in this research? Note that if the LLM is used only for writing, editing, or formatting purposes and does \emph{not} impact the core methodology, scientific rigor, or originality of the research, declaration is not required.
    \item[] Answer: \answerNA{}
    \item[] Justification: LLMs are not used as a component of the core methodology. The evaluated MLLMs are the subject of the benchmark study, not a methodological tool.

\end{enumerate}

\end{document}